\documentclass[journal]{IEEEtran}
\usepackage{amsmath,amsfonts}
\usepackage{algorithmic}
\usepackage{array}
\usepackage{textcomp}
\usepackage{stfloats}
\usepackage{url}
\usepackage{verbatim}
\usepackage{graphicx}
\usepackage{xcolor}
\newcommand{\rev}[1]{#1}
\def\BibTeX{{\rm B\kern-.05em{\sc i\kern-.025em b}\kern-.08em
    T\kern-.1667em\lower.7ex\hbox{E}\kern-.125emX}}
\usepackage{balance}
\usepackage{amsmath}
\usepackage{amssymb}
\usepackage{enumerate}
\usepackage{enumitem}
\usepackage[caption=false,font=footnotesize,labelfont=rm,textfont=rm]{subfig}

\begin{document}
\title{StableMotion: One-Step Motion Estimation with Diffusion Prior}
\author{Ziyi~Wang,
    Haipeng~Li,
    Lin~Sui,
    Tianhao~Zhou,
    Hai~Jiang,
    Lang~Nie, \\
    Bing Zeng,~\IEEEmembership{Fellow,~IEEE},
    Shuaicheng~Liu,~\IEEEmembership{Senior Member,~IEEE}%
    \thanks{Ziyi Wang and Tianhao Zhou are with the Yingcai \rev{Honors} College, University of Electronic Science and Technology of China, Chengdu 611731, China (email: \{ziyiwang, thzhou\}@std.uestc.edu.cn).}%
    \thanks{Haipeng Li, Bing Zeng and Shuaicheng Liu are with the School of Information and Communication Engineering, University of Electronic Science and Technology of China, Chengdu 611731, China (email: \{lihaipeng@std., eezeng@, liushuaicheng@\}uestc.edu.cn).}%
    \thanks{Lin Sui is with 4Paradigm Inc., Beijing 100080, China (email: suilin0432@gmail.com).}%
    \thanks{Hai Jiang is with the School of \rev{Aeronautics and Astronautics}, Sichuan University, Chengdu 610065, China (email: jianghai@stu.scu.edu.cn).}%
    \thanks{Lang Nie is with the School of Artificial Intelligence, Chongqing University of Posts and Telecommunications, Chongqing 400065, China (\rev{email}: nielang@cqupt.edu.cn).}%
    \thanks{This work was supported by National Natural Science Foundation of China under Grant No.62372091.}%
    \thanks{\textit{Corresponding author: Shuaicheng Liu}}%
}


\maketitle

\begin{abstract}

    We present StableMotion, a novel framework that leverages geometric and content priors from pretrained large-scale image diffusion models for motion estimation in single-image rectification tasks such as Stitched Image Rectangling (SIR) and Rolling Shutter Correction (RSC). Specifically, StableMotion takes a text-to-image Stable Diffusion (SD) model as its backbone and repurposes it as an image-to-motion estimator. To mitigate inconsistent outputs produced by diffusion models, we propose Adaptive Ensemble Strategy (AES), which consolidates multiple outputs into a cohesive, high-fidelity result.
    Additionally, we present Sampling Steps Disaster (SSD), a counterintuitive phenomenon in which increasing the number of sampling steps can lead to poorer outcomes, motivating our one-step inference design. StableMotion is evaluated on two image rectification tasks and delivers state-of-the-art performance on both, while also showing {promising transferability through qualitative examples and no-reference evaluations on unseen SIR-OOD and real-captured RSC benchmarks}. Supported by SSD, StableMotion {achieves efficient one-step inference, offering over 100$\times$ speedup compared to previous diffusion model-based methods even when combined with the optional AES post-processing}. Code and weights are available at \url{https://github.com/ivowang/StableMotion}.
\end{abstract}


\section{Introduction}
\begin{figure}
    \begin{center}
        \includegraphics[width=1\linewidth]{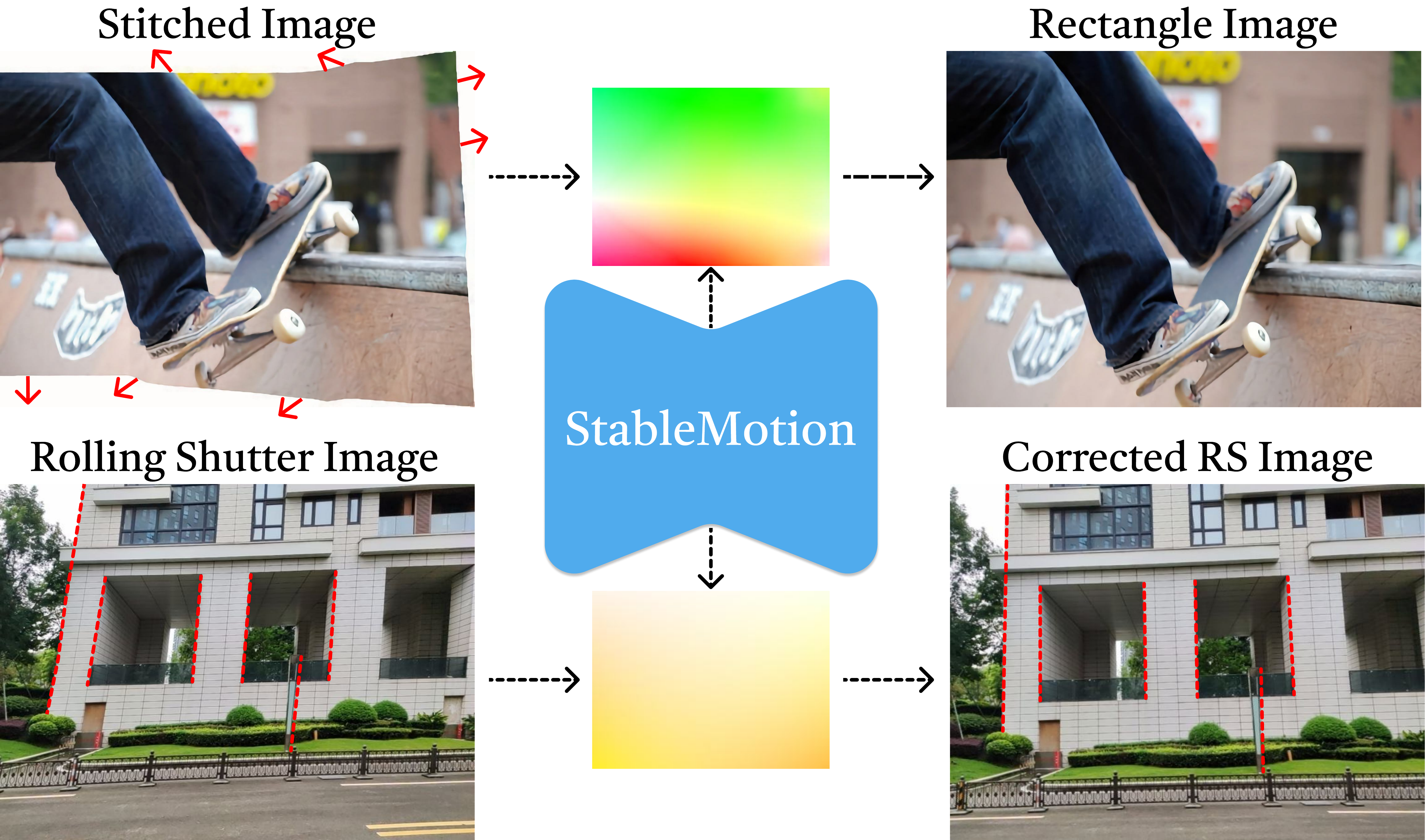}
    \end{center}
    \caption{Two applications of the StableMotion framework. The upper part shows the application of StableMotion on image rectangling, while the lower part shows the application of StableMotion on rolling shutter correction. StableMotion accomplishes different tasks by generating corresponding rectification flows.}
    \label{fig:teaser}
\end{figure}
Single-image-based image rectification tasks, such as Stitched Image Rectangling (SIR) and Rolling Shutter Correction (RSC), are fundamental computer vision tasks with broad applications in various domains. SIR is essential for creating visually appealing images from stitched photos, widely used in image stitching~\cite{brown2007automatic} and panorama construction \cite{brown2003recognising}. RSC addresses the distortion artifacts caused by CMOS sensors' row-wise exposure pattern, which is crucial for high-quality video recording \cite{7457352} and 3D reconstruction \cite{7780732, 6751167}. These tasks are particularly challenging due to their ill-posed nature, where the information available in the input image is insufficient to obtain robust results, necessitating additional prior knowledge or constraints.

Stitched Image Rectangling (SIR) refers to the technique of converting stitched images with irregular edges into rectangular shapes. These stitched images are typically created by merging multiple overlapping images~\cite{szeliski2007image,zaragoza2013projective,lee2020warping,jiang2024towards,nie2021unsupervised}.
\cite{he2013rectangling} proposed the concept of image rectangling and presented an initial framework. However, it primarily preserves straight lines and can introduce distortions in non-linear structures~\cite{zhu2022semi}.
In the deep-learning era, \cite{nie2022deep} proposed a mesh-based framework. However, meshes contain far fewer grid points than pixels, and this low-rank representation cannot fully capture complex motions, resulting in local artifacts such as inconsistent boundaries or visible misalignments. More recently, \cite{zhou2023rectangling} introduced a specially designed diffusion-based solution. Because it trains two diffusion models from scratch and performs diffusion in pixel space, both training and inference are computationally intensive. Moreover, the performance of these methods is strongly tied to the scale and quality of the task-specific datasets, which limits further improvement.

Rolling Shutter Correction (RSC) aims to rectify image distortions arising from the row-wise exposure pattern of CMOS sensors. For single-frame rolling shutter correction, \cite{UnrollingShutter_CVPR17} used CNN modules to estimate row-wise motion, but did not handle cases involving camera or scene movement beyond the horizontal direction well. \cite{Zhuang_2019_CVPR} introduced depth maps as an additional input, but depth estimation is itself ill-posed. \cite{Yan_2023_ICCV} used a homography mixture model, and \cite{yang2024single} proposed a dedicated diffusion model. Both approaches are trained from scratch and still rely heavily on high-quality datasets and substantial computation, which limits their potential and makes generalization challenging.

Meanwhile, diffusion models have rapidly emerged as a transformative force in generative modeling. With their ability to synthesize high-quality, diverse, and semantically meaningful content, models such as Stable Diffusion (SD)~\cite{rombach2022high} have set new standards in image generation, restoration, and related tasks. Beyond image synthesis, these models also show a strong capacity to capture semantic and structural information \cite{tang2023emergent,luo2024diffusion,zhang2024tale,hedlin2024unsupervised}. We attribute this capability to the rich knowledge embedded in large-scale diffusion models, which enables them to model complex relationships and transformations in visual data.

Motivated by this insight, we harness diffusion priors for the challenging problem of single-image motion estimation. Specifically, we introduce \textbf{StableMotion}, a general framework that repurposes the architecture and pretrained weights of Stable Diffusion (as illustrated in Fig.~\ref{fig:teaser}) for motion estimation in image rectification tasks. StableMotion adapts to different inputs, predicts the corresponding rectification flows, and applies to multiple tasks in this domain. Our motivation is also supported by prior work that leverages SD priors for tasks such as image restoration~\cite{luo2024taming} and depth estimation~\cite{ke2024repurposing}. However, these methods are \textit{image-to-image} frameworks and therefore suffer from unstable content generation and slow inference. Meanwhile, SD has also shown the ability to perceive motion between different images~\cite{tang2023emergent,luo2024diffusion,zhang2024tale,hedlin2024unsupervised}, enabling zero-shot semantic and geometric matching. To unlock this potential for motion perception, we formulate StableMotion as an \textit{image-to-motion} model. A repurposed VAE maps images and motions between pixel space and latent space, while the UNet is adapted to estimate motion fields \( F \) between the conditional input images \( I_{cond} \) (i.e., stitched images for SIR and rolling shutter images for RSC) and the corresponding ground-truth images \( I_{gt} \) (i.e., rectangular images for SIR and corrected rolling shutter images for RSC). The estimated motion fields are then used to warp \( I_{cond} \) and produce the predicted image \( \hat I_{gt} \).

However, since the initial noise is randomly sampled from Gaussian distributions and randomness is introduced during denoising, diffusion models can produce multiple inconsistent results when performing inference on the same input. To address this issue, we introduce \textbf{Adaptive Ensemble Strategy (AES)} as an optional post-processing step to aggregate inconsistent results into a unified output.

{Beyond the inconsistency issue, we identify a failure mode of conditional diffusion models trained with auxiliary losses. We find that introducing conditional losses can lead to a counterintuitive phenomenon: using more sampling steps can degrade performance. This observation challenges the conventional view that more sampling steps necessarily improve diffusion-model outputs. We introduce \textbf{Sampling Steps Disaster (SSD)} as a local perturbation analysis for this phenomenon. SSD supports our single-step inference design, improving efficiency in practice, and also provides a mechanistic explanation for the empirical choice of using fewer sampling steps in previous works~\cite{li2024dmhomo,zhou2023rectangling,yang2024single}.}

In sum, StableMotion delivers state-of-the-art performance while substantially reducing training and inference cost, and also shows {promising transfer to unseen data, supported by qualitative examples and complementary no-reference evaluations on unseen SIR-OOD and real-captured RSC benchmarks}.
Our contributions are:
\begin{itemize}
    \item We present \textbf{StableMotion}, a novel framework that repurposes a foundation model for motion estimation, unlocking {its} semantic and structural priors. We demonstrate its effectiveness on two image rectification tasks.
    \item We present the concept of \textbf{Sampling Steps Disaster (SSD)}, which {helps explain} the paradox that increasing the number of sampling steps can lead to poorer outcomes and supports the one-step inference design of StableMotion.
    \item Extensive experiments demonstrate that StableMotion achieves state-of-the-art performance on public benchmarks of both tasks, shows {promising transferability on unseen data}, and markedly reduces training and inference cost, {offering over 100$\times$ speedup compared to previous DM-based methods (32\,ms per forward pass; approximately 64\,ms with default AES ensemble of $n{=}2$)}.
\end{itemize}

\begin{figure*}[t]
    \begin{center}
        \includegraphics[width=1\linewidth]{figures_/Group_733.pdf}
    \end{center}
    \caption{{Repurposing from SD. Taking image rectangling as an example, the conditions (i.e., stitched images and their mask) and motion fields (scaled by the hyperparameter $\gamma$) are encoded into feature space. The flow features are noised and concatenated with the condition features, and then fed into the denoising network. At each timestep, the predicted flow feature is decoded and denormalized into pixel space to perform warping, constructing the conditional loss.
            }}
    \label{fig:training_ppl}
\end{figure*}

\section{Related Work}
\subsection{Diffusion Models}
Diffusion models have emerged as powerful generative models~\cite{sohl2015deep,ho2020denoising}. They can also be formulated within score-based frameworks~\cite{song2019generative, song2021scorebased}, which focus on estimating gradients of the data distribution. Techniques such as classifier-guided diffusion~\cite{dhariwal2021diffusion, liu2023more} and classifier-free guidance~\cite{ho2021classifierfree} provide finer control over the generation process by incorporating auxiliary information or adjusting guidance strength. Built on these ideas, Latent Diffusion Models (LDMs)~\cite{rombach2022high} perform diffusion in latent space and thereby improve efficiency. In addition, approaches including ControlNet~\cite{zhang2023adding} and related methods~\cite{wang2022zero, hu2022lora,  karras2022elucidating, lu2022dpm, hoogeboom2023simple, li2024dmhomo, jiang2023low} leverage pretrained diffusion priors to adapt the generative process to specific tasks.
Diffusion models have also been widely applied to low-level vision tasks such as super-resolution~\cite{luo2024skipdiff, yuan2024efficient} and depth estimation~\cite{ke2024repurposing}.

\subsection{Prior-Based Methods}
Foundational models such as Stable Diffusion~\cite{rombach2022high} and DeepFloyd~\cite{saharia2022photorealistic} encode extensive high-level semantic information and have become useful for many downstream tasks. They are used in different ways: some methods extract features from these models for semantic or geometric matching~\cite{tang2023emergent,luo2024diffusion,zhang2024tale,hedlin2024unsupervised}, visual anagrams~\cite{geng2024visual,chen2024images}, 3D reconstruction~\cite{poole2022dreamfusion, wu2024reconfusion}, image segmentation~\cite{xiao2023text,wang2023diffusion}, and classification~\cite{li2023your}. Other methods directly adapt these models for tasks such as depth estimation~\cite{ke2024repurposing} and 3D geometry estimation~\cite{fu2024geowizard}. In this work, we exploit diffusion-based image priors for motion estimation.

\subsection{Motion Estimation Tasks}
Motion estimation determines the geometric transformation, or motion, between images. Reliable estimates support numerous applications, including image or video stitching \cite{nie2021unsupervised,7563334}, video stabilization \cite{727336,10.1145/2461912.2461995,7563334,7909025}, layout rectangling \cite{nie2022deep,zhou2023rectangling}, rolling-shutter correction \cite{yang2024single}, image rotation \cite{he2013content,10128955}, and fisheye-distortion removal \cite{1642666}. These transformations are typically formulated as warping operations described by pixel-wise flows~\cite{10.1007/978-3-540-24673-2_3}, mesh flows~\cite{10.1007/978-3-319-46466-4_48}, or homographies~\cite{DBLP:journals/corr/DeToneMR16}. In this work, we introduce StableMotion, a versatile framework that estimates pixel-wise motion fields, and demonstrate its effectiveness on two applications: stitched image rectangling (SIR) and rolling-shutter correction (RSC).

Stitched image rectangling transforms an image produced by image stitching, which often has irregular non-rectangular boundaries, into a regular rectangular image. The problem was first posed in \cite{he2013rectangling}, which proposed a two-stage framework: the first stage initializes an irregular mesh through seam carving~\cite{avidan2023seam}, and the second stage solves for a content-preserving rectangular mesh by energy optimization. Building on this idea, \cite{nie2022deep} pioneered a deep-learning approach to image rectangling by estimating mesh deformations with convolutional neural networks (CNNs). More recently, \cite{zhou2023rectangling} employed specially designed diffusion models for image rectangling using a two-stage cascade involving \textit{image-to-motion} and \textit{image-to-image} transformations. \cite{zhang2024recstitchnet} designed an end-to-end framework that combines image stitching and rectangling, and \cite{nie2024semi} proposed a semi-supervised rectangling method.

Rolling shutter correction aims to mitigate or eliminate geometric distortions and artifacts introduced by rolling shutter cameras. For single-frame RSC, classical algorithms usually use lines and boundaries in RS images to estimate motion~\cite{rengarajan_2016_cvpr, lao_2018_cvpr}. In the deep-learning era, \cite{UnrollingShutter_CVPR17} employed CNN modules to generate row-wise motion between global-shutter (GS) images and rolling-shutter (RS) inputs. \cite{Zhuang_2019_CVPR} added depth maps as an additional input, although depth estimation is itself ill-posed. \cite{Yan_2023_ICCV} used a homography mixture model that divides images into blocks and learns coefficients to combine several motion bases. More recently, \cite{yang2024single} proposed the first diffusion-based method for single-image RSC, using a dedicated diffusion model to estimate motion in pixel space.

\section{Method}
\subsection{Overview}

We adopt the priors (i.e., pretrained weights) from existing foundation models, specifically Stable Diffusion 2.0~\cite{rombach2022high}, as our backbone. The model is expected to predict a flow field $F \in \mathbb{R}^{2 \times H \times W}$ representing the motion of each pixel from the condition image $I_{cond}$ towards the corresponding ground truth image ${I}_{gt}$:
\begin{equation}
    \label{eq:sd_predict_flow}
    \hat{F} = \theta_{} (C, \epsilon),
\end{equation}
where $\epsilon$ is standard Gaussian noise, and $C$ denotes the conditioning input, which at minimum contains the input image $I_{cond}$. For SIR, $C = (I_{cond}, M)$, where $I_{cond}$ denotes the stitched image and $M$ is a mask indicating the blank regions in $I_{cond}$. For RSC, $C = I_{cond}$, where $I_{cond}$ denotes the rolling-shutter image. The final prediction is obtained through a warping operation:
\begin{equation}
    \label{eq:warp_IS}
    \hat{I}_{gt} = \mathcal{W}(I_{cond}, \hat{F}).
\end{equation}

\subsection{Preliminaries}

\begin{figure*}[t]
    \begin{center}
        \includegraphics[width=1\linewidth]{figures_/Group_732.pdf}
    \end{center}
    \caption{Overview of the inference scheme, using image rectangling as an example. We perform one-step inference based on the understanding of Sampling Steps Disaster, and use Adaptive Ensemble Strategy to address generative instability.}
    \label{fig:sample_ppl}
\end{figure*}

\paragraph{Classifier-free guidance}
To control the content generated by the model and balance controllability with fidelity, classifier-free guidance (CFG)~\cite{ho2021classifierfree} incorporates conditions $\mathbf{y}$ as follows:
\begin{equation}
    p_\theta(\mathbf{x}_{t-1} | \mathbf{x}_t, \mathbf{y}) = \mathcal{N}\left(\mathbf{x}_{t-1}; \mu_\theta\left(\mathbf{x}_t, t, \mathbf{y}\right), \sigma_t^2\mathbf{I}\right).
    \label{dm:p_sing_cond}
\end{equation}
CFG theoretically requires two diffusion outputs, one conditional and one unconditional. Joint training is usually applied by randomly setting the condition $y$ to a null condition $\phi$ with some probability $p_{y}$, and \rev{the} sampling process is a linear combination of conditional and unconditional predictions:
\begin{equation}
    \tilde{\mu}_{\theta}\left(\mathbf{z}_{\lambda}, \mathbf{y}\right)=(1+w) \mu_{\theta}\left(\mathbf{z}_{\lambda}, \mathbf{y}\right) - w \mu_{\theta}\left(\mathbf{z}_{\lambda}, \phi\right),
\end{equation}
where $w$ is the parameter that balances fidelity and diversity. In our work, we use full-condition guidance, which means $p_{y}$ and $w$ are both set to zero.

\paragraph{Latent diffusion models}
Our methodology is built upon the framework of Latent Diffusion Models (LDMs)~\cite{rombach2022high}, which adapt the diffusion process to a compressed latent space to significantly enhance computational efficiency. While standard Diffusion Models (DMs)~\cite{sohl2015deep,ho2020denoising} have demonstrated exceptional performance in image synthesis, their iterative denoising process in the high-dimensional pixel space incurs substantial computational costs. LDMs circumvent this issue by first employing a \rev{pretrained} VAE, consisting of an encoder $\mathcal{E}$ and a decoder $\mathcal{D}$, to perform transformations between pixel space and feature space:
\begin{equation}
    z^{(x)} = \mathcal{E}(x), \quad x \approx \mathcal{D}(z^{(x)}).
    \label{eq:LDM_encode}
\end{equation}
The latent space is designed to preserve perceptually relevant information in a compact representation, improving computational efficiency while retaining useful semantic structure.

\subsection{Repurposing SD for Motion Estimation}
We adapt the UNet into a motion estimator while retaining the pretrained VAE as a fixed latent representation, as illustrated in Fig.~\ref{fig:training_ppl}.

\paragraph{VAE adaptation}

To enable the flow fields to be compatible with the input range of the \rev{pretrained} VAE, we introduce a normalization step. Specifically, we define a scaling factor $\gamma$ that approximates the maximum absolute value within the original flow map. This scalar ensures that the normalized flow values fall within the range expected by the VAE, typically $[-1, 1]$. Denoting the original flow as $F$ and the normalized counterpart as $f$, the normalization and subsequent denormalization processes are defined as follows:
\begin{equation}
    f = \frac{F}{\gamma}, \quad \hat{F} = \hat{f} \cdot \gamma,
    \label{eq:norm2}
\end{equation}
where $\hat{f}$ denotes the reconstructed normalized flow output by the VAE decoder, and $\hat{F}$ is the recovered flow in the original scale.

To ensure geometric consistency under affine transformations, we further augment the normalized flow $f \in \mathbb{R}^{H \times W \times 2}$ by appending an all-ones channel to form a homogeneous coordinate representation:
\begin{equation}
    f^{\prime} = \text{Concat}(f, \mathbf{1}),
    \label{eq:input_cat}
\end{equation}
where $\mathbf{1} \in \mathbb{R}^{H \times W \times 1}$ is a constant channel filled with ones, and $f^{\prime} \in \mathbb{R}^{H \times W \times 3}$ is the resulting homogeneous flow field.

Both the conditioning input $C$ (e.g., structural cues or guidance images) and the homogeneous flow $f^{\prime}$ are encoded into the latent space via a shared VAE encoder $\mathcal{E}(\cdot)$:
\begin{equation}
    z^{(C)} = \mathcal{E}(C), \quad z^{(f^{\prime})} = \mathcal{E}(f^{\prime}),
\end{equation}
where $z^{(C)}$ and $z^{(f^{\prime})}$ denote the latent representations of the condition and the normalized flow respectively. These latent codes are subsequently used for conditional sampling and reconstruction during the diffusion-based generation process.

\paragraph{UNet adaptation}
The input to the UNet module, denoted by $z_{\text{in}}$, is the concatenation of the latent condition and latent flow features:
\begin{equation}
    z_{\text{in}} = \text{cat}[z^{(C)}, z_{t}^{(f^{\prime})}],
\end{equation}
where $z_{t}^{(f^{\prime})}$ represents $z^{(f^{\prime})}$ after $t$ steps of forward diffusion. The original SD UNet accepts a 4-channel input. In our setting, however, $z_{\text{in}}$ contains $4(N+1)$ channels, where $N$ denotes the number of condition elements. For SIR, the conditions $C$ include the stitched image and its mask, giving $N_{SIR} = 2$. For RSC, the condition $C$ is the rolling-shutter image, giving $N_{RSC} = 1$. For each task, we replicate and combine the UNet input layer accordingly, and scale the resulting weights by a factor of $N+1$ to preserve a reasonable initialization.

The UNet output is $\hat{z}_0^{(f\prime)}$, which is decoded into the homogeneous flow $\hat{f}^\prime$. We then take the first two channels of $\hat{f}^\prime$ to form the normalized flow $\hat{f}$. After denormalization, we obtain the predicted flow $\hat{F}$.

\paragraph{Training}
\label{sec:loss}
{We initialize the framework with pretrained Stable Diffusion 2.0 weights. During training, the pretrained VAE encoder and decoder are kept frozen, and only the UNet is updated.} The training loss is a convex combination of several loss items. For each timestep $t$, the estimated flow feature is given by:
\begin{equation}
    \label{eq:v2f}
    \hat{z}^{(f^\prime)}_{0|t} = \alpha_t z^{(f^\prime)}_t - \sigma_t \hat{\mu}_{\theta}(z^{(C)}, z^{(f^\prime)}_t),
\end{equation}
where $\alpha_t$ and $\sigma_t$ are the forward diffusion parameters. Based on that, loss functions are constructed as follows:

Firstly, diffusion reconstruction loss is given by:
\begin{equation}
    \label{eq:loss_mse}
    \ell_{diff} = \left\| z^{(f^\prime)} - \hat{z}^{(f^\prime)}_{0|t} \right\|_2.
\end{equation}
Secondly, we impose a condition loss in pixel space. Let $C$ denote the conditions, $C_{gt}$ the corresponding ground truth, and $\hat{F}_{0|t}$ the predicted flow at timestep $t$ after denormalization. The condition loss is defined as:
\begin{equation}
    \label{eq:loss_cond}
    \ell_{cond} = \left\| C_{gt} - \mathcal{W}(C, \hat{F}_{0|t}) \right\|_2.
\end{equation}

Additionally, we compute a perceptual loss using a pretrained VGG-16 model $v_{\theta}$~\cite{johnson2016perceptual} to improve visual fidelity. Let $I_{gt}$ denote the ground-truth image (e.g., the rectangular image for SIR and the global-shutter image for RSC). The perceptual loss is given by:
\begin{equation}
    \label{eq:loss_pct}
    \ell_{pct} = \left\| v_{\theta}(I_{gt}) - v_{\theta}(\mathcal{W}(I_{cond}, \hat{F}_{0|t})) \right\|_2.
\end{equation}

The training loss is a weighted sum of them, that is:
\begin{equation}
    \label{eq:loss_total}
    \ell_{total} = \ell_{diff} + \lambda_1 \ell_{cond} + \lambda_2 \ell_{pct},
\end{equation}
where $\lambda_1$ and $\lambda_2$ are hyperparameters that balance the contributions of each loss term.

Prior work such as~\cite{zhou2023rectangling} uses a two-stage diffusion-based framework, with one diffusion model for motion generation and another for post-hoc image refinement. In contrast, StableMotion predicts motion fields directly with a single adapted UNet and decodes them through the frozen pretrained VAE. This keeps the framework compact and avoids an additional image-generation refinement stage. Empirically, this design reduces boundary artifacts while preserving the efficiency of one-step inference, as shown in Section~\ref{sec:qualt}.

\paragraph{Inference}
Fig.~\ref{fig:sample_ppl} illustrates the overall inference pipeline. The input conditions $C$ are encoded into latent representations using the VAE encoder, producing $z^{(C)}$. We also initialize a latent variable $z_T^{(f')}$ by sampling from a standard Gaussian distribution. This sampled noise is concatenated with the condition latent $z^{(C)}$ and fed into the denoising UNet. A single DDIM~\cite{song2020denoising} step is then applied as an approximation of $p(z_0^{(f')}|z_T^{(f')})$. The predicted latent feature $\hat z_0^{(f')}$ is decoded by the VAE into the flow field $\hat F$ in pixel space. Finally, the estimated flow $\hat F$ is warped onto the condition image $I_{cond}$ to produce the rectified output image $\hat I_{gt}$. The rationale for using one-step inference is discussed in Section~\ref{sec:ssd}.
\begin{figure}[t]
    \centering

    \subfloat[Training with condition loss. The black and green arrow corresponds to the supervision of $\rev{\ell_{diff}}$ and $\rev{\ell_{cond}}$, pointing to the distribution of pseudo flow and ground truth flow, respectively. The red arrow is the joint effect of them, pointing to the learned distribution $y_0$. ]{
        \includegraphics[width=0.96\linewidth]{figures_/Group_488.pdf}
        \label{fig:ssd1}
    }

    \subfloat[Inference with SSD. Given the model prediction $y_{0|t}$, the noise scheduler calculates the input of the next sampling step $y_{t-1}$. Because $y_t$ \rev{follows a different distribution from} $x_t$, directly using $y_t$ as the condition in the next step yields \rev{an} error. Such an error accumulates with the increase of sampling steps, namely Sampling Steps Disaster.]{
        \includegraphics[width=0.96\linewidth]{figures_/Frame_26.pdf}
        \label{fig:ssd2}
    }

    \caption{Explanation of Sampling Steps Disaster (SSD). When multiple target distributions are involved, directly performing multi-step inference introduces accumulating error.}
    \label{fig:ssd}
\end{figure}

\subsection{Sampling Steps Disaster}
\label{sec:ssd}

In addition to the diffusion reconstruction loss, training diffusion models with additional losses, such as the condition and perceptual losses used in our framework, has proven effective for improving model performance~\cite{li2024dmhomo, zhou2023rectangling, yang2024single}. However, once such auxiliary losses are introduced, increasing the number of sampling steps can lead to worse inference performance, which is counterintuitive.

We take the condition loss as \rev{an} example to investigate this issue and illustrate our analysis in Fig.~\ref{fig:ssd}. We use the symbols \(x_t\) and \(z_t\) to represent elements from the Pseudo Label Distribution at timestep \(t\) (\(PD_t\)) and the Ground Truth Distribution at the same timestep (\(GD_t\)), respectively. The loss functions \(\ell_{diff}\) and \(\ell_{cond}\) guide the model \(\theta\) to learn the conditional distributions \(p(x_0|x_{t})\) and \(p(z_0|x_{t})\), shown by the black and green dashed lines in Fig.~\ref{fig:ssd1}. Under their joint constraint, the conditional distribution learned by our model \(\theta\) is intermediate. We use the symbol \(y_0\) to represent elements in the Learned Distribution (\(LD_t\)) between \(PD_t\) and \(GD_t\), with the conditional distribution \(p(y_0|x_{t})\) indicated by the red arrow in Fig.~\ref{fig:ssd1}.

Fig.~\ref{fig:ssd2} illustrates the inference process with multiple steps. In the first step, the model takes pure noise \(x_T\) as input and generates an output \(y_{0|T}\). The key issue arises in the second step. Instead of receiving \(x_{T-1}\), which is the kind of input seen during training, the model receives \(y_{T-1}\). Notably, \(y_{T-1}\) {follows a different distribution from} \(x_{T-1}\). Thus, the inference chain is disrupted from the second step onward, as shown by the gray dashed arrow with a cross in Fig.~\ref{fig:ssd2}. This error can accumulate as sampling continues. {We now formalize this observation.}

{In diffusion-based inference, each sampling step applies the denoiser $\theta$ followed by the noise scheduler $s_t$ to produce the next intermediate state. When the model is trained with auxiliary losses (e.g., condition loss, perceptual loss), the denoiser's learned output distribution ($LD$) may deviate from the pseudo-label distribution ($PD$) that provides the forward-process noisy inputs. Consequently, after the first step, the denoiser can receive intermediate states that are biased away from the training-time input distribution. We formalize this effect as a local perturbation analysis: the per-step distribution gap is modeled by small correction vectors, and the analysis studies how these corrections propagate through the sampling chain.}

{\textbf{Assumption 1} (Local smoothness). \textit{Let $p_t:\mathbb{R}^d\to\mathbb{R}^d$ denote the composite mapping $s_t\circ\theta$ (scheduler $\circ$ denoiser) at timestep $t$. For the local analysis below, assume each $p_t$ is twice continuously differentiable ($C^2$) on a neighborhood of the relevant trajectories, with Jacobian matrix $J_{p_t}(z)\in\mathbb{R}^{d\times d}$ at point $z$. In particular, each $J_{p_t}$ is locally Lipschitz, which ensures the Taylor remainder is $O(\|h\|^2)$.}}

{This assumption should be read as a standard local linearization condition. In practice, the SD UNet is composed mostly of smooth operations (e.g., SiLU, normalization, attention) and the DDIM scheduler is affine in the denoiser output, so this approximation is appropriate in the neighborhoods considered by the analysis.}

{\textbf{Definition 1} (Local distribution-gap correction). \textit{Let $\bar{y}_T=x_T$ and $\bar{y}_{t-1}=p_t(\bar{y}_t)$ be the uncorrected inference trajectory, i.e., the trajectory produced by feeding the model's own intermediate states back into the next denoising step. For each $t<T$, let $\Delta_t$ denote a small local correction that moves the current state of the corrected trajectory toward the training-time pseudo-label input distribution $PD_t$. We set $\Delta_T=0$ because the initial state is pure Gaussian noise shared by the forward process and inference.}}

{Intuitively, $\Delta_t$ measures the local mismatch between an intermediate state produced by the learned sampling trajectory and the type of noisy input on which the denoiser was trained at timestep $t$. We do not require a unique pointwise pairing between $PD_t$ and $LD_t$; only the local gap magnitude and its propagation enter the expansion.}

{\textbf{Definition 2} (Inference processes). \textit{Starting from $x_T$ (pure noise, shared across distributions):}}
\begin{itemize}[leftmargin=2em,nosep]
    \item {\textit{Uncorrected process.} $\bar{y}_{t-1}=p_t(\bar{y}_t)$ for $t=T,\ldots,1$, with $\bar{y}_T=x_T$. Output: $\bar{y}_0$.}
    \item {\textit{Corrected process.} $\hat{y}_{T-1}=p_T(x_T)=\bar{y}_{T-1}$; then for $t=T{-}1,\ldots,1$: $\hat{x}_t=\hat{y}_t+\Delta_t$, $\hat{y}_{t-1}=p_t(\hat{x}_t)$. Output: $\hat{x}_0=\hat{y}_0+\Delta_0$.}
\end{itemize}
{The uncorrected process is what actually happens during multi-step inference. The corrected process is a thought experiment that injects a local distribution-gap correction before each subsequent denoising step. Their difference $e_t=\hat{x}_t - \bar{y}_t$ measures the first-order effect of such per-step mismatches on the trajectory.}

{\textbf{Notation.} For $a\leq b$, define the ordered Jacobian product}
\begin{equation}
    {A_{a:b}\triangleq J_{p_a}(\bar{y}_a)\,J_{p_{a+1}}(\bar{y}_{a+1})\cdots J_{p_b}(\bar{y}_b),}
    \label{eq:jacobian_product}
\end{equation}
{with the convention $A_{a:b}=I$ (identity) when $a>b$. This product captures how a perturbation at step $b$ is amplified as it propagates through the denoising chain from step $b$ back to step $a$. All norms $\|\cdot\|$ below denote the Euclidean vector norm and its induced operator norm.}

{\textbf{Lemma 1} (Error recursion). \textit{Define the per-step error $e_t\triangleq\hat{x}_t - \bar{y}_t$. Then $e_{T-1}=\Delta_{T-1}$ (base case), and for $t\leq T{-}2$:}
\begin{equation}
    {e_t = J_{p_{t+1}}(\bar{y}_{t+1})\cdot e_{t+1}+\Delta_t + r_t, \quad \|r_t\|\leq C_{t+1}\|e_{t+1}\|^2,}
    \label{eq:error_recursion}
\end{equation}
{\textit{where $C_{t+1}$ is a constant depending on the local curvature of $p_{t+1}$.}}}

{Lemma~1 states that the error at step $t$ consists of two parts: (i) the local distribution gap $\Delta_t$ newly introduced at this step, and (ii) the previous error $e_{t+1}$ amplified by the Jacobian $J_{p_{t+1}}$, i.e., the local sensitivity of the denoiser to input perturbations.}

{\textit{Proof.} \textbf{Base case.} At $t=T{-}1$, both processes share the same first step: $\hat{y}_{T-1}=p_T(x_T)=\bar{y}_{T-1}$. By the corrected-process definition, $\hat{x}_{T-1}=\hat{y}_{T-1}+\Delta_{T-1}$, so}
\begin{equation}
    {e_{T-1} = \hat{x}_{T-1}-\bar{y}_{T-1} = \Delta_{T-1}.}
    \label{eq:base_case}
\end{equation}

{\textbf{Inductive step.} For $t\leq T{-}2$, suppose $e_{t+1}$ is known. By definition, $\hat{x}_{t+1}=\bar{y}_{t+1}+e_{t+1}$. Applying $p_{t+1}$:}
\begin{equation}
    {\hat{y}_t = p_{t+1}\!\left(\hat{x}_{t+1}\right) = p_{t+1}\!\left(\bar{y}_{t+1}+e_{t+1}\right).}
    \label{eq:lemma_step1}
\end{equation}
{Since $p_{t+1}\in C^2$ (Assumption~1), the Taylor expansion with Lagrange remainder gives:}
\begin{equation}
    {
        \begin{aligned}
            \hat{y}_t
            &= p_{t+1}(\bar{y}_{t+1})
            + J_{p_{t+1}}(\bar{y}_{t+1})\cdot e_{t+1}
            + r_t',\\
            \|r_t'\| &\leq C_{t+1}\|e_{t+1}\|^2 .
        \end{aligned}
    }
    \label{eq:lemma_taylor}
\end{equation}
{where $C_{t+1}\triangleq\frac{1}{2}\sup_{\xi}\|H_{p_{t+1}}(\xi)\|_{\mathrm{op}}$ over $\xi$ on the segment between $\bar{y}_{t+1}$ and $\hat{x}_{t+1}$, and $H_{p_{t+1}}$ denotes the Hessian. Noting that $\bar{y}_t = p_{t+1}(\bar{y}_{t+1})$ by the uncorrected process:}
\begin{equation}
    {\hat{y}_t - \bar{y}_t = J_{p_{t+1}}(\bar{y}_{t+1})\cdot e_{t+1} + r_t'.}
    \label{eq:lemma_diff}
\end{equation}
{Adding the correction $\Delta_t$ via $e_t = \hat{x}_t - \bar{y}_t = (\hat{y}_t - \bar{y}_t) + \Delta_t$:}
\begin{equation}
    {e_t = J_{p_{t+1}}(\bar{y}_{t+1})\cdot e_{t+1} + \Delta_t + r_t', \quad \|r_t'\|\leq C_{t+1}\|e_{t+1}\|^2.  \hfill\square}
    \label{eq:lemma_result}
\end{equation}

{By unrolling the recursion in Lemma~1 across all timesteps, we can express the total output error as a sum over all intermediate gaps $\Delta_k$, each amplified by the cumulative Jacobian product from that step to the output. This is analogous to how gradients propagate through layers in a deep network, where each layer's sensitivity multiplies into the total.}

{\textbf{Theorem 1} (Sampling Steps Disaster). \textit{Under Assumption~1, for any fixed finite number of inference steps $T$, let $\eta\triangleq\max_{0\leq t\leq T-1}\|\Delta_t\|$. Then the local expansion of the output error is:}
\begin{equation}
    {e_0 = \Delta_0 + \sum_{k=1}^{T-1}A_{1:k}\,\Delta_k + R_T, \quad \|R_T\|=O_T(\eta^2),}
    \label{eq:ssd_theorem}
\end{equation}
{\textit{where the ordered Jacobian product $A_{1:k}$ is defined in~\eqref{eq:jacobian_product}. The notation $O_T(\eta^2)$ means that the hidden constant may depend on $T$ and on local bounds of the Jacobians and Hessians along the relevant trajectories, but not on $\eta$.}}}

{\textit{Proof.} We proceed by induction on $t$ (from $T{-}1$ down to $0$). We prove the general form: for each $0\leq t\leq T{-}1$,}
\begin{equation}
    {e_t = \Delta_t + \sum_{k=t+1}^{T-1}A_{t+1:k}\,\Delta_k + R_t, \quad \|R_t\|=O_T(\eta^2).}
    \label{eq:thm_general}
\end{equation}

{\textbf{Base case} ($t=T{-}1$). From Lemma~1, $e_{T-1}=\Delta_{T-1}$. Since the sum $\sum_{k=T}^{T-1}(\cdot)$ is empty:}
\begin{equation}
    {e_{T-1} = \Delta_{T-1} + 0 + 0,}
    \label{eq:thm_base}
\end{equation}
{which matches~\eqref{eq:thm_general} with $R_{T-1}=0$.}

{\textbf{Inductive hypothesis.} Assume~\eqref{eq:thm_general} holds for $t+1$, i.e.:}
\begin{equation}
    {
        \begin{aligned}
            e_{t+1}
            &= \Delta_{t+1}
            + \sum_{k=t+2}^{T-1}A_{t+2:k}\,\Delta_k
            + R_{t+1},\\
            \|R_{t+1}\| &= O_T(\eta^2).
        \end{aligned}
    }
    \label{eq:thm_ih}
\end{equation}

{\textbf{Inductive step.} Substituting~\eqref{eq:thm_ih} into the recursion~\eqref{eq:error_recursion} (dropping the remainder $r_t$ temporarily):}
\begin{equation}
    {e_t = J_{p_{t+1}}(\bar{y}_{t+1})\cdot e_{t+1} + \Delta_t + r_t.}
    \label{eq:thm_sub}
\end{equation}
{Expanding $J_{p_{t+1}}(\bar{y}_{t+1})\cdot e_{t+1}$ using~\eqref{eq:thm_ih}:}
\begin{equation}
    {
        \begin{aligned}
            J_{p_{t+1}}(\bar{y}_{t+1})\cdot e_{t+1}
            &= A_{t+1:t+1}\,\Delta_{t+1}
            + \sum_{k=t+2}^{T-1}A_{t+1:k}\,\Delta_k \\
            &\quad + J_{p_{t+1}}(\bar{y}_{t+1})\,R_{t+1}.
        \end{aligned}
    }
    \label{eq:thm_expand}
\end{equation}
{Merging $A_{t+1:t+1}\,\Delta_{t+1}$ into the summation (extending it to start from $k=t{+}1$):}
\begin{equation}
    {
        \begin{aligned}
            e_t
            &= \Delta_t
            + \sum_{k=t+1}^{T-1}A_{t+1:k}\,\Delta_k \\
            &\quad + \underbrace{J_{p_{t+1}}(\bar{y}_{t+1})\,R_{t+1}
            + r_t}_{\triangleq\, R_t}.
        \end{aligned}
    }
    \label{eq:thm_result}
\end{equation}
{For fixed finite $T$, the local Jacobian products in the first-order term are bounded on the neighborhoods considered. Hence the inductive hypothesis implies $e_{t+1}=O_T(\eta)$, so $\|r_t\|\leq C_{t+1}\|e_{t+1}\|^2=O_T(\eta^2)$. Multiplication by the locally bounded Jacobian $J_{p_{t+1}}$ preserves $O_T(\eta^2)$, and therefore $\|R_t\|=O_T(\eta^2)$, completing the induction.}

{Setting $t=0$ in~\eqref{eq:thm_result} yields~\eqref{eq:ssd_theorem}. \hfill $\square$}

{\textbf{Corollary 1} (Error bound). \textit{If $\|J_{p_t}(\bar{y}_t)\|_{\mathrm{op}}\leq L$ for all $t$ and $\|\Delta_t\|\leq\delta$ for all $t$, then the first-order term satisfies:}
\begin{equation}
    {\left\|\Delta_0 + \sum_{k=1}^{T-1}A_{1:k}\,\Delta_k\right\|\leq\delta\!\sum_{k=0}^{T-1}L^k\!=\!\begin{cases}\delta\cdot\frac{L^T\!-\!1}{L\!-\!1}, & \!\!L\!\neq\! 1,\\ T\delta, & \!\!L\!=\!1.\end{cases}}
    \label{eq:ssd_bound}
\end{equation}
{\textit{When $L>1$, this upper bound grows exponentially with $T$, indicating that the linearized error can be exponentially amplified as the number of inference steps increases. This is an upper bound on the first-order term, not a claim that every trajectory grows monotonically.}}}

{This condition should be interpreted locally. We do not claim that every denoiser has $L>1$ or that the error must increase for every sample. Rather, when auxiliary losses create nonzero per-step gaps $\Delta_t$ and the Jacobian products along a trajectory have norm larger than one, intermediate mismatches can be amplified through the sampling chain. This is the regime captured by SSD.}

{\textbf{Corollary 2} (No accumulation at $T{=}1$). \textit{When $T=1$, the summation $\sum_{k=1}^{0}(\cdot)$ is empty, so $e_0=\Delta_0+R_T$ with $\|R_T\|=O_T(\eta^2)$. The multi-step accumulation terms vanish entirely, leaving only the single-step output gap $\Delta_0$. This does not prove global optimality of one-step inference; it shows that one-step inference removes the intermediate mismatch-propagation terms.}}

{\textbf{Remark} (Connection to standard diffusion models). In the idealized case of a standard diffusion model trained only with the denoising score-matching loss $\ell_{diff}$ and optimized perfectly, the learned denoising trajectory matches the pseudo-label forward-process distribution, so the auxiliary-loss-induced gaps $\Delta_t$ vanish. In practice, ordinary model error and discretization error may still exist; SSD refers specifically to the additional mismatch introduced when auxiliary losses (e.g., $\ell_{cond}$, $\ell_{pct}$) pull the learned output distribution away from $PD$. This perspective explains why prior diffusion-based methods~\cite{li2024dmhomo, zhou2023rectangling, yang2024single} that incorporate such losses empirically observed better results with fewer steps, but could not explain the phenomenon theoretically. Our analysis supports single-step inference because it applies the denoiser once from the shared pure-noise input and avoids propagation through intermediate mismatch terms.}

For models with multiple training objectives, such as those {that incorporate} conditional losses, this phenomenon is commonly observed. Table~\ref{table:step_compare_supp} shows the PSNR with different steps sampled from MDM~\cite{zhou2023rectangling} with a DDIM~\cite{song2020denoising} scheduler, where the phenomenon of Sampling Steps Disaster has also emerged.

\begin{table}[htb]
    \centering
    \def\temptablewidth{0.46\textwidth}
    {\rule{\temptablewidth}{1pt}}
    \begin{tabular*}{\temptablewidth}{@{\extracolsep{\fill}}c|ccccc}
        \hline
        Inference steps &1& 4 & 16 & 64& 256 \\
        \hline
        PSNR & {22.14} & {21.84} & {21.43} & {21.31} & {21.26}\\
        \hline
    \end{tabular*}
    {\rule{\temptablewidth}{1pt}}
    \caption{Sampling Steps Disaster has also emerged in other works, such as MDM~\cite{zhou2023rectangling}.}
    \label{table:step_compare_supp}
\end{table}

\begin{figure}[thb]
    \begin{center}
        \includegraphics[width=\linewidth]{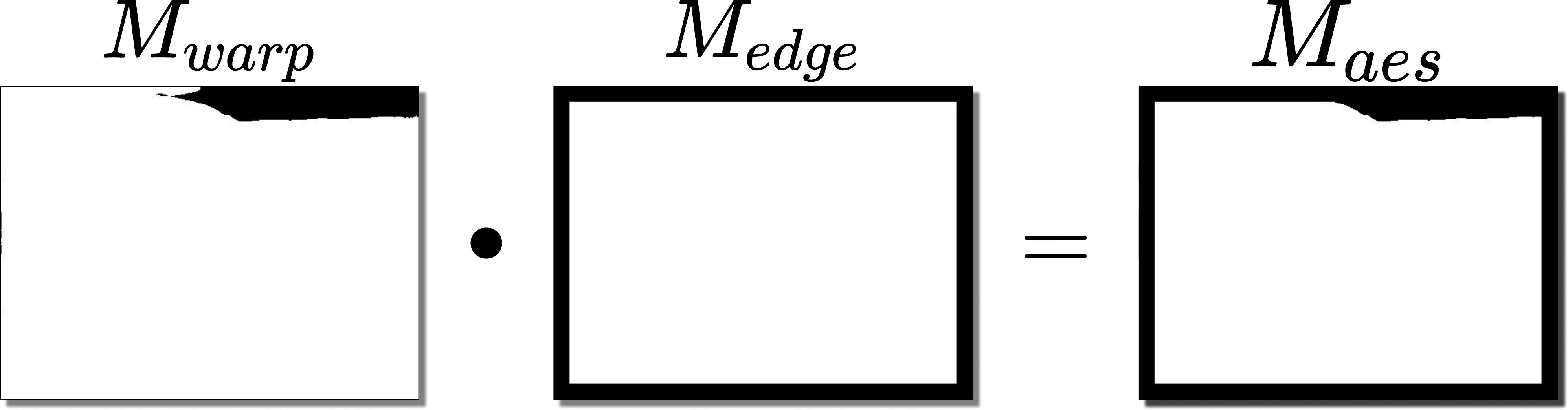}
    \end{center}
    \caption{Calculation of $M_{aes}$.}
    \label{fig:AES}
\end{figure}
\subsection{Adaptive Ensemble Strategy} 
\begin{figure}[thb]
    \begin{center}
        \includegraphics[width=\linewidth]{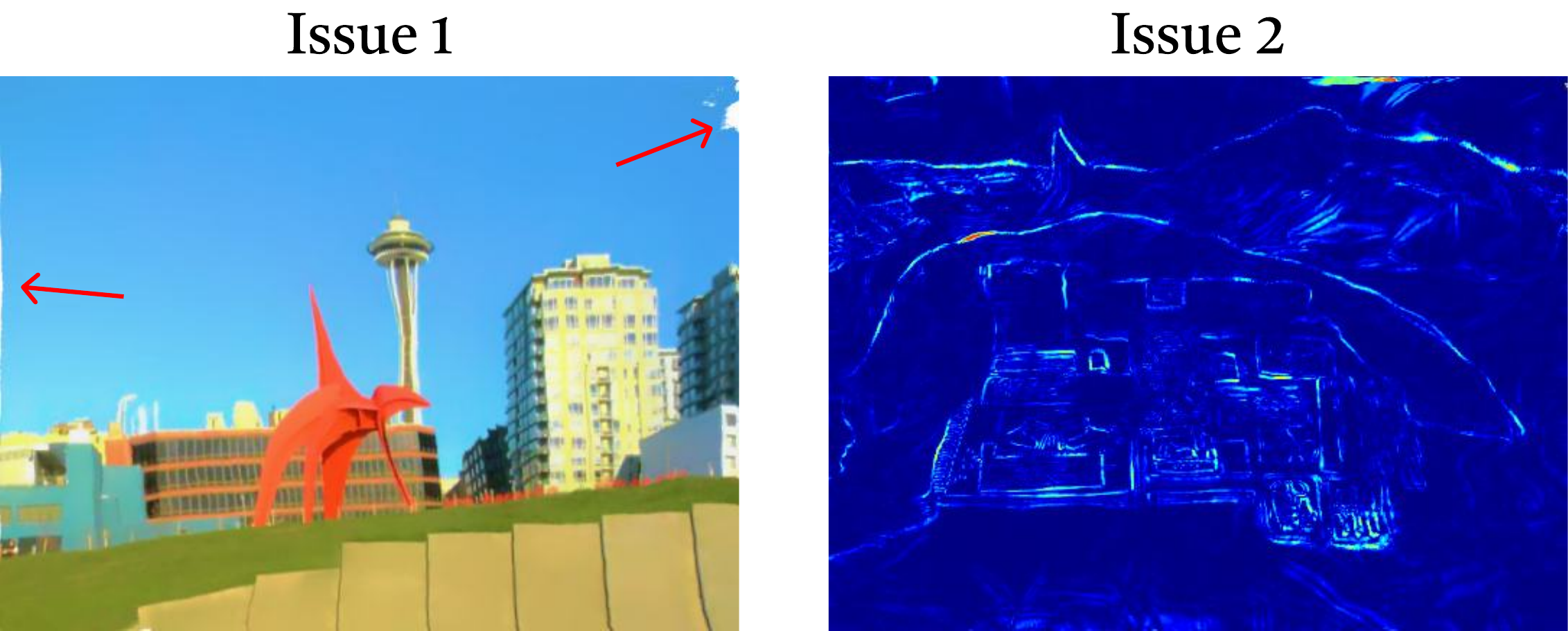}
    \end{center}
    \caption{Illustration of the two main issues addressed by AES. \textbf{Left:} The red arrow highlights the white margin artifacts that commonly appear along the boundaries after warping, representing the first issue. \textbf{Right:} The heatmap visualizes the pixel-wise differences between multiple outputs generated from the same input. Bright regions indicate inconsistency among the samples, revealing the generative instability of diffusion models and motivating the need for an ensemble strategy to improve output consistency.}
    \label{fig:AES_Issue}
\end{figure}

Adaptive Ensemble Strategy (AES) is a post-process step designed to solve the following issues, as shown in Fig.~\ref{fig:AES_Issue}:
\begin{description}[
        leftmargin=4.3em,
        align=left,
        labelsep=1em
    ]
    \item [Issue 1.] In tasks such as image rectangling, images are warped according to a motion field, which can create boundary artifacts, typically blank white margins around the warped image.
    \item [Issue 2.] Diffusion models can produce inconsistent results due to their generative nature, leading to variations in the output images.
\end{description}
The first issue is localized to the margin areas and can be addressed with a minimum-filter ensemble: pixels in the white margins have the highest intensities, so taking the minimum value across samples tends to suppress white borders. However, applying such a filter everywhere can blur edges and degrade image quality. The second issue affects the whole image, where a median filter can reduce noise and increase consistency, but cannot resolve the boundary issue by itself. Therefore, we adopt distinct ensemble strategies for different regions, which we call Adaptive Ensemble Strategy (AES).

This strategy is {built} on an adaptive mask $M_{aes}$, which indicates existing or potential margins. According to this mask, we apply different filters to different regions: pixels marked as ``possible margins'' are processed with a minimum filter, while the remaining pixels are processed with a median filter. This design aggregates multiple outputs into a unified final image and addresses both issues simultaneously.

{Specifically, for SIR, $M_{aes}$ is computed as the element-wise product of two masks.} The first mask \(M_{warp}\) is defined as follows:
\begin{equation}
    \label{eq:M21}
    M_{warp} = \text{warp}(M, \hat{F}_{0|t}).
\end{equation}
{It is an adaptive mask revealing existing margins in the warped image. The second mask, $M_{edge}$, uses fixed edges to ensure that potential margins are not missed because of the uncertainty of $\hat{F}_{0|t}$. Through element-wise multiplication, the final mask $M_{aes}$ marks the combined margin region retained by both masks:}
\begin{equation}
    \label{eq:M2}
    M_{aes} = M_{warp} \cdot M_{edge}.
\end{equation}
{For RSC, AES is applied by using the median filter on the valid regions marked by $M_{mask}$, where $M_{mask}$ denotes the binary mask of valid (non-margin) regions in the RSC output, i.e., pixels that are not affected by rolling-shutter boundary artifacts.}

\begin{figure*}[htb]
    \begin{center}
        \includegraphics[width=\linewidth]{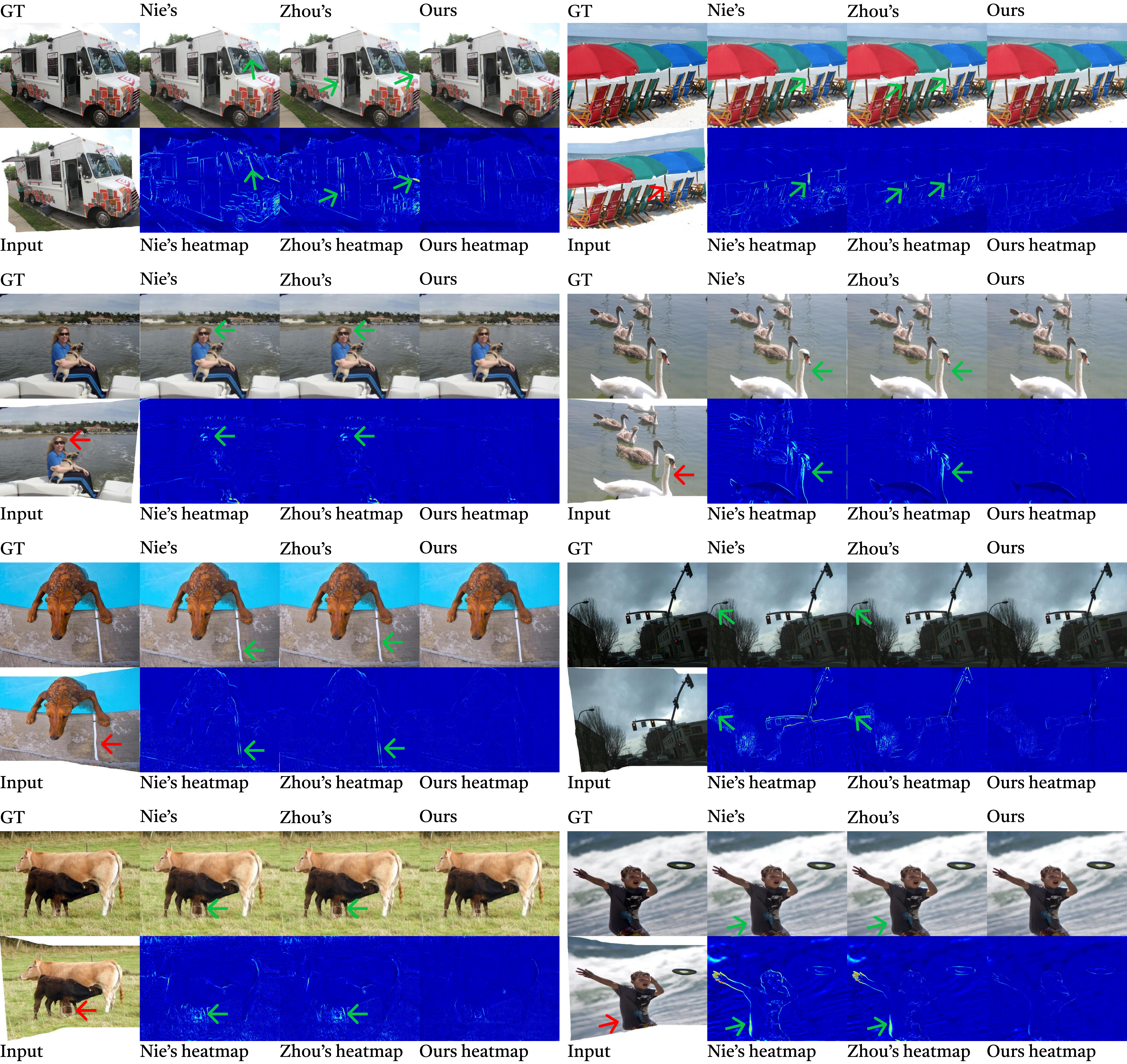}
    \end{center}
    \caption{Comparison of image rectangling with previous methods. Green arrows point to artifacts or unresolved margins. The brighter areas on the heatmap indicate a greater discrepancy between the output and the ground truth.}
    \label{fig:qualitative_compare_sir}
\end{figure*}

\section{Experiment}

\begin{figure*}[t]
    \begin{center}
        \includegraphics[width=\linewidth]{figures_/Group_658.pdf}
    \end{center}
    \caption{Comparison of rolling shutter correction with previous methods. The brighter areas on the heatmap indicate a greater discrepancy between the output and the ground truth.}
    \label{fig:qualitative_compare_rsc}
\end{figure*}

\begin{figure*}[htb]
    \begin{center}
        \includegraphics[width=\linewidth]{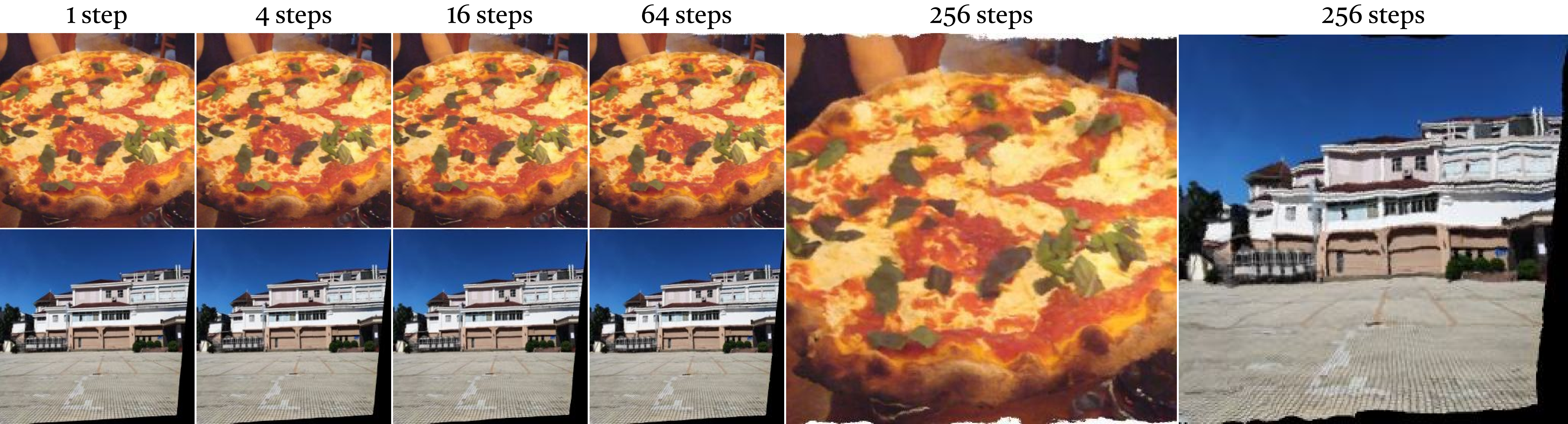}
    \end{center}
    \vspace{-1em}
    \caption{Visual comparison of different sampling steps with SSD.}
    \label{fig:sampling_steps_quali}
\end{figure*}

\begin{figure*}[t]
    \begin{center}
        \includegraphics[width=\linewidth]{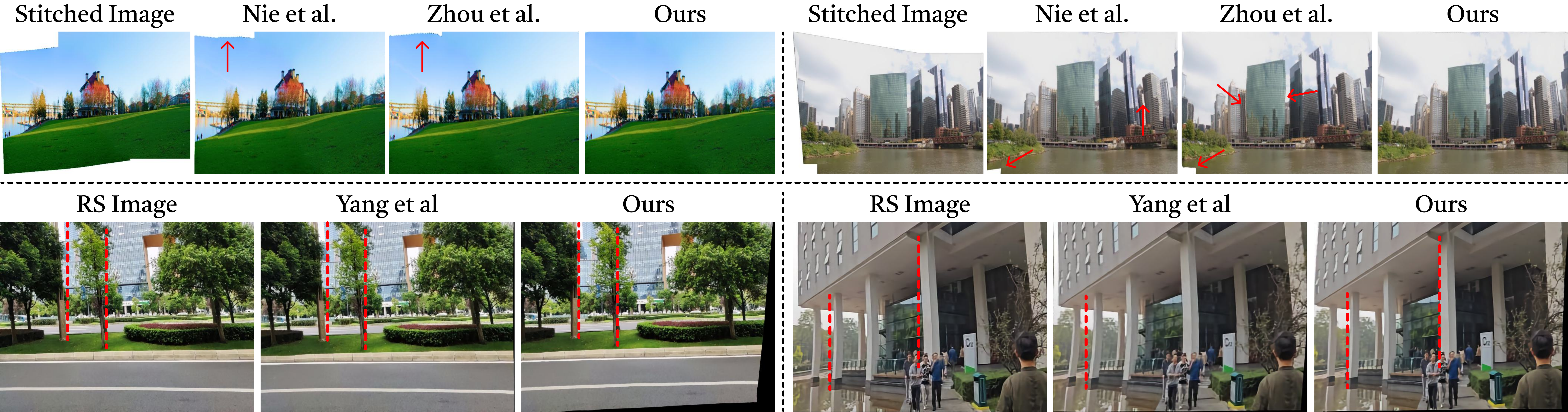}
    \end{center}
    \caption{{Qualitative zero-shot transfer examples.} StableMotion shows promising transfer behavior in addressing irregular boundaries and rolling shutter distortions on unseen data.}
    \label{fig:gene}
\end{figure*}

\subsection{Implementation Details\label{sec:imple}}
StableMotion takes Stable Diffusion 2.0~\cite{rombach2022high} as backbone and employs a 1000-step DDPM scheduler~\cite{ho2020denoising} for training. The pretrained VAE encoder and decoder are kept frozen, while only the UNet is fine-tuned. The learning rate was set to $2 \times 10^{-5}$, with batch sizes of 128 for SIR and 32 for RSC. Training takes 60,000 steps, spanning 10 hours. {During training, each iteration uniformly samples a timestep from the 1000-step DDPM schedule; StableMotion is not trained with a one-step distillation target. At inference, it uses a single DDIM step from the initial noise state to the predicted clean flow latent.} {For inference, a single forward pass takes 32\,ms on one NVIDIA H100. With the default AES setting of $n{=}2$ ensemble samples, the total inference time is approximately 64\,ms per image (the AES filtering post-processing adds negligible overhead). We note that our main quantitative results (Tables~\ref{table:IR_metrics} and \ref{table:RSC_metrics}) are reported with $n{=}2$ by default.} $\lambda_1$ and $\lambda_2$ are set to 1 and 0.01, respectively. For training, we use the DIR-D dataset~\cite{nie2022deep} for SIR and the RS-Real dataset~\cite{yang2024single} for RSC.

\subsection{Quantitative Comparison\label{sec:quant}}
To comprehensively evaluate the quality of the generated and corrected images, we employ a combination of distortion-based and perception-based metrics. This allows for a balanced assessment, capturing both pixel-level accuracy and similarity to human visual perception: distortion metrics (PSNR and SSIM) measure the absolute difference between a reference image and a generated image, and perceptual metrics (LPIPS and FID) utilize deep learning models to assess image quality in a way that is more consistent with human judgment. For image rectangling, we include traditional method~\cite{he2013rectangling}, the first deep-learning-based approach~\cite{nie2022deep}, the latest semi-supervised method CoupledTPS~\cite{nie2024semi}, and the latest diffusion model-based method RecDiffusion~\cite{zhou2023rectangling}. For rolling shutter correction, we include the homography mixture method~\cite{Yan_2023_ICCV} and the diffusion-based method~\cite{yang2024single}. As shown in Table~\ref{table:IR_metrics} and Table~\ref{table:RSC_metrics}, {StableMotion} achieves state-of-the-art performance in most categories.

\begin{figure}[htb]
    \centering
    \includegraphics[width=0.95\linewidth]{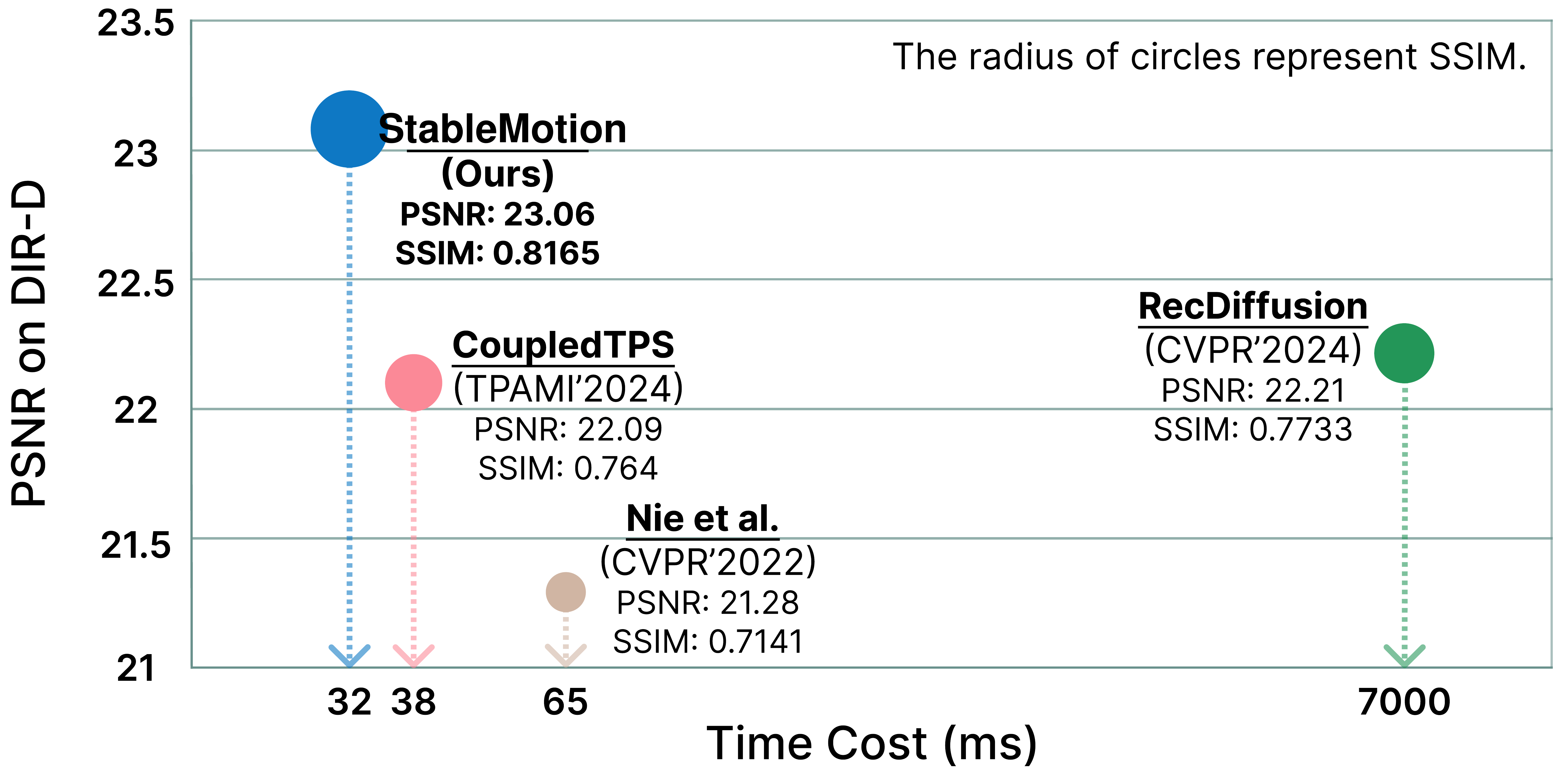}
    \caption{Comparison of inference time cost of different methods.}
    \label{fig:time_cost}
\end{figure}

Specifically, {StableMotion} not only generates results closer to the ground truth, improving metrics related to data consistency, but also delivers stronger perceptual quality by leveraging semantic priors from the foundation model.
The inference {times} of these methods are {reported} in Fig.~\ref{fig:time_cost}.

\subsection{Qualitative Comparison\label{sec:qualt}}
\begin{table}[htb]
    \centering
    \def\temptablewidth{0.49\textwidth}
    {\rule{\temptablewidth}{1pt}}
    \begin{tabular*}{\temptablewidth}{@{\extracolsep{\fill}}l|cccc}
        \hline
        Method & PSNR$\uparrow$ & SSIM$\uparrow$ & LPIPS$\downarrow$ & FID$\downarrow$ \\
        \hline
        He et al.~\cite{he2013rectangling} & {14.70} & {0.378} & {-} & 38.19\\
        Nie et al.~\cite{nie2022deep} & {21.28} & {0.714} & {0.152} & 21.77 \\
        CoupledTPS \cite{nie2024semi} & {22.09} & {0.764} & {0.140} & 20.02 \\
        RecDiffusion \cite{zhou2023rectangling} & {22.21} & {0.773} & {0.789} & 18.75 \\
        \hline
        StableMotion (Ours) & \textbf{23.06} & \textbf{0.817} &\textbf{0.136} &\textbf{17.22} \\
        \hline
    \end{tabular*}
    {\rule{\temptablewidth}{1pt}}
    \caption{Quantitative comparison of different methods on DIR-D testing set~\cite{nie2022deep}. {Best scores are highlighted in \textbf{bold}.}}
    \label{table:IR_metrics}
\end{table}

\begin{table}[htb]
    \centering
    \def\temptablewidth{0.49\textwidth}
    {\rule{\temptablewidth}{1pt}}
    \begin{tabular*}{\temptablewidth}{@{\extracolsep{\fill}}l|cccc}
        \hline
        Method & PSNR$\uparrow$ & SSIM$\uparrow$ & LPIPS$\downarrow$ & FID$\downarrow$ \\
        \hline
        Yan et al.~\cite{Yan_2023_ICCV} & {18.76} & {0.549} & {0.117} & 8.37 \\
        RS-Diffusion~\cite{yang2024single} & {22.02} & {0.704}  & \textbf{0.067} & 5.68  \\
        \hline
        StableMotion (Ours) & \textbf{22.65} & \textbf{0.724} &{0.068} &\textbf{5.18}\\
        \hline
    \end{tabular*}
    {\rule{\temptablewidth}{1pt}}
    \caption{Quantitative comparison of different methods on RS-Real testing set~\cite{yang2024single}. {Best scores are highlighted in \textbf{bold}.}}
    \label{table:RSC_metrics}
\end{table}

Our method is evaluated against the previous state-of-the-art methods on DIR-D~\cite{nie2022deep} for SIR and on RS-Real~\cite{yang2024single} for RSC, as shown in Fig.~\ref{fig:qualitative_compare_sir} and Fig.~\ref{fig:qualitative_compare_rsc}.

For SIR, we compare our results specifically with those of \cite{nie2022deep} and \cite{zhou2023rectangling}.
Previous methods are prone to introducing local distortions that are visually unappealing.
Results of \cite{nie2022deep} are often misaligned, leaving visible local seams, as indicated by the green arrow in the figure. \cite{zhou2023rectangling} easily distorts both linear and non-linear structures, like the twisted car door in the first row.
StableMotion better preserves both linear and non-linear structures, reduces white boundary artifacts, and alleviates distortions in the input image.

For RSC, StableMotion also produces results closer to the ground truth, as reflected in significantly fewer bright spots in the \rev{alignment} heatmap.

We also observe an interesting phenomenon in some cases. Previous stitching methods rely on warping operations to combine multiple images, which can introduce distortions in the stitched input, as indicated by the red arrows in Fig.~\ref{fig:qualitative_compare_sir} (Input). Existing rectification methods often preserve these distortions. In contrast, StableMotion can alleviate some of them, which we conjecture is related to the semantic and structural priors carried by the pretrained model and decoder.

\subsection{Zero-Shot Transfer and No-Reference Evaluation}

To provide qualitative evidence of transferability, we perform zero-shot inference on the APAP-Conssite dataset~\cite{zaragoza2013projective} using the model trained on DIR-D~\cite{nie2022deep} for image rectangling (above the dotted line), and on newly captured RS images using the model trained on RS-Real~\cite{yang2024single} for rolling shutter correction (below the dotted line). As illustrated in Fig.~\ref{fig:gene}, the method performs competitively relative to previous methods in several cases, particularly in handling white edges and linear structures on unseen data.

{To further reduce the possibility of cherry-picked qualitative evidence, we evaluate SIR generalization on the cross-scenario benchmark from RopStitch~\cite{nie2026robust}, which is primarily sourced from~\cite{lin2015adaptive,chang2014shape,chen2016natural,li2017parallax,herrmann2018object}, and RSC generalization on two real-captured benchmarks from RS-Diffusion~\cite{yang2024single}, which consist of rolling-shutter video frames and real rolling-shutter photographs. These benchmarks do not provide paired ground-truth rectified images, so reference-based metrics such as PSNR and SSIM cannot be computed. We therefore use task-specific no-reference GPT-5.4 scoring protocols as complementary evidence. For SIR, the evaluator scores rectangular boundary, content preservation, geometry/structure, artifact control, and naturalness with weights 0.25, 0.25, 0.25, 0.15, and 0.10. For RSC, the evaluator scores RS suppression, geometry/structure, content preservation, artifact control, and naturalness with weights 0.30, 0.25, 0.20, 0.15, and 0.10. The same prompt and weights are used for all methods within each task; the exact SIR and RSC prompts are reported in black boxed prompt panels. The per-sample rubric scores are integers from 0 to 10, and Table~\ref{table:ood_score} reports benchmark-level averages. \rev{We query the pinned model version \texttt{gpt-5.4} through the OpenAI Responses API (evaluations conducted in May 2026) with high-detail image input and strict JSON-schema decoding. We do not override the sampling parameters, so the model runs with the API's default temperature and top-$p$, while the prompt, rubric weights, and per-task reasoning effort are kept fixed across all methods. Within every request the input image is presented first and a candidate output second; moreover, each method is scored in a \emph{separate} request that sees only the input and that single method's output, so competing methods are never co-presented and no left/right or presentation-order bias arises across methods, and per-pair preferences are derived offline from the independently assigned overall scores. We regard this no-reference evaluation only as complementary supporting evidence, not as a replacement for reference-based metrics (e.g., PSNR and SSIM) or controlled human studies, and we therefore read the out-of-distribution results as indicative rather than conclusive.}}

\begin{figure}[!t]
\centering
{\color{black}
\setlength{\fboxsep}{5pt}
\setlength{\fboxrule}{0.4pt}
\fcolorbox{black}{white}{%
\begin{minipage}{0.92\columnwidth}
\scriptsize
\raggedright
\textbf{Prompt.}

\vspace{0.04cm}
You are an expert reviewer for stitched image rectangling (SIR).

\vspace{0.04cm}
Task definition:

\vspace{0.04cm}
- The input image is a stitched image with irregular boundaries, white/empty regions, and possibly geometric deformation.

\vspace{0.04cm}
- The output image is the candidate rectangling result. A strong result should form a clean rectangular image while preserving the input content and field of view, keeping scene geometry natural, avoiding line discontinuities/local distortions, suppressing white borders and warping artifacts, and not hallucinating unrelated new content.

\vspace{0.04cm}
Score only what can be inferred by comparing the two provided images. Penalize outputs that crop away large valid content, add unrelated objects, leave white/empty borders, bend salient structures, break continuous lines, duplicate content, blur details, or introduce seams/noise.

\vspace{0.04cm}
Return JSON only. Use integer scores from 0 to 10 for each rubric dimension: (1) rectangular\_boundary: Is the output a complete, clean rectangle without visible empty/white/invalid borders? (2) content\_preservation: Does the output preserve the input scene content and field of view without excessive cropping or unrelated hallucination? (3) geometry\_structure: Are salient geometry, straight lines, object shapes, and non-linear structures visually plausible and continuous? (4) artifact\_control: Are warping artifacts, seams, discontinuities, blur, noise, duplicated regions, and local corruption avoided? (5) naturalness: Does the final image look like a coherent natural photograph?

\vspace{0.04cm}
Use the full 0--10 range. A score of 10 is near publication-quality; 7--8 is good with minor issues; 4--6 has clear but not catastrophic problems; 1--3 is poor; 0 is unusable or impossible to judge.
\end{minipage}}
}
\caption{{The scoring prompt for no-reference evaluation of SIR generalization.}}
\label{fig:prompt}
\end{figure}

\begin{figure}[!t]
\centering
{\color{black}
\setlength{\fboxsep}{5pt}
\setlength{\fboxrule}{0.4pt}
\fcolorbox{black}{white}{%
\begin{minipage}{0.92\columnwidth}
\scriptsize
\raggedright
\textbf{Prompt.}

\vspace{0.04cm}
You are an expert reviewer for single-image rolling shutter correction (RSC). 

\vspace{0.04cm}
Task definition:

\vspace{0.04cm}
- The input is a rolling-shutter (RS) image captured with row-wise exposure, so straight structures may lean, bend, wobble, or curve, and scene content may be skewed by camera/device motion.

\vspace{0.04cm}
- \rev{The output is one candidate corrected image.} A strong output should look closer to a global-shutter (GS) image: reduced row-wise skew/wobble, straighter buildings/poles/edges, plausible object shapes, preserved scene content, and no over-correction or new artifacts.

\vspace{0.04cm}
\rev{Score only what can be inferred from the two provided images: INPUT and OUTPUT. Do not compare against another method, because no other method output is visible in this conversation.} Penalize residual RS distortion, over-correction, bent straight lines, stretched local regions, ghosting, tearing, duplicated content, blur/noise, severe crop, hallucinated or missing objects, and unnatural appearance.

\vspace{0.04cm}
Return JSON only. Use integer scores from 0 to 10 for \rev{the output}: (1) rs\_suppression: How well are rolling-shutter skew, wobble, bending, and row-wise deformation corrected? (2) geometry\_structure: Are salient straight lines, buildings, poles, road edges, object shapes, and moving objects geometrically plausible and continuous? (3) content\_preservation: Does the output keep the input scene content and field of view without excessive cropping, missing regions, or hallucination? (4) artifact\_control: Are local warping artifacts, tearing, seams, ghosting, blur, duplicated content, and corruption avoided? (5) naturalness: Does the corrected image look like a coherent natural global-shutter photograph?

\vspace{0.04cm}
Use the full 0--10 range. A score of 10 is near publication-quality; 7--8 is good with minor issues; 4--6 has clear but not catastrophic problems; 1--3 is poor; 0 is unusable or impossible to judge.
\end{minipage}}
}
\caption{{The scoring prompt for no-reference evaluation of RSC generalization.}}
\label{fig:rsc_prompt}
\end{figure}

{As shown in Table~\ref{table:ood_score}, StableMotion obtains a higher overall score than RecDiffusion~\cite{zhou2023rectangling} on the SIR-OOD benchmark from RopStitch~\cite{nie2026robust} (6.86 vs.\ 6.31). On the RS-Diffusion~\cite{yang2024single} RSC video-frame benchmark, StableMotion improves the overall score from 6.12 to 6.51; on the RS-Diffusion real-photo benchmark, it improves the overall score from 6.76 to 7.29. The RSC gains mainly come from stronger RS suppression and more plausible geometry, while content preservation is slightly lower than RS-Diffusion. These results provide complementary no-reference evidence of \rev{promising} transfer to unseen SIR-OOD and real-captured RSC cases.}

\begin{table}[tb]
    \centering
    \def\temptablewidth{0.49\textwidth}
    {\color{black}
        {\rule{\temptablewidth}{1pt}}
        \scriptsize
        \begin{tabular*}{\temptablewidth}{@{\extracolsep{\fill}}llcccccc}
            \hline
            Data & Method & Overall & Spec. & Geom. & Cont. & Art. & Nat.\\
            \hline
            SIR-OOD & RecDiff. & 6.31 & 7.63 & 4.91 & 7.57 & 4.99 & 5.35\\
            SIR-OOD & Ours & \textbf{6.86} & \textbf{8.84} & \textbf{5.18} & \textbf{7.74} & \textbf{5.60} & \textbf{5.78}\\
            RSC-F & RS-Diff. & 6.12 & 5.75 & 5.91 & \textbf{6.59} & 6.44 & 6.35\\
            RSC-F & Ours & \textbf{6.51} & \textbf{6.46} & \textbf{6.72} & 6.21 & \textbf{6.58} & \textbf{6.67}\\
            RSC-P & RS-Diff. & 6.76 & 6.06 & 6.17 & \textbf{8.06} & \textbf{7.33} & 6.89\\
            RSC-P & Ours & \textbf{7.29} & \textbf{7.06} & \textbf{7.06} & 7.89 & \textbf{7.33} & \textbf{7.33}\\
            \hline
        \end{tabular*}
        {\rule{\temptablewidth}{1pt}}
    }
    \caption{{No-reference generalization evaluation on unseen benchmarks. Scores are assigned by GPT-5.4 from 0 to 10 and averaged over each benchmark; higher is better. The SIR-OOD benchmark is from RopStitch~\cite{nie2026robust}; the SIR baseline is RecDiffusion~\cite{zhou2023rectangling}. RSC-F and RSC-P denote real RS video frames and real RS photographs from RS-Diffusion~\cite{yang2024single}, respectively. \rev{Spec.\ denotes the task-specific primary dimension (rectangular boundary quality for SIR-OOD, and RS suppression for RSC-F/RSC-P); Geom., Cont., Art., and Nat.\ denote geometry/structure, content preservation, artifact control, and naturalness.}}}
    \label{table:ood_score}
\end{table}

\subsection{Sampling Steps Disaster}
\label{sec:exp_ssd}
\begin{figure}[htb]
    \centering
    \includegraphics[width=\linewidth]{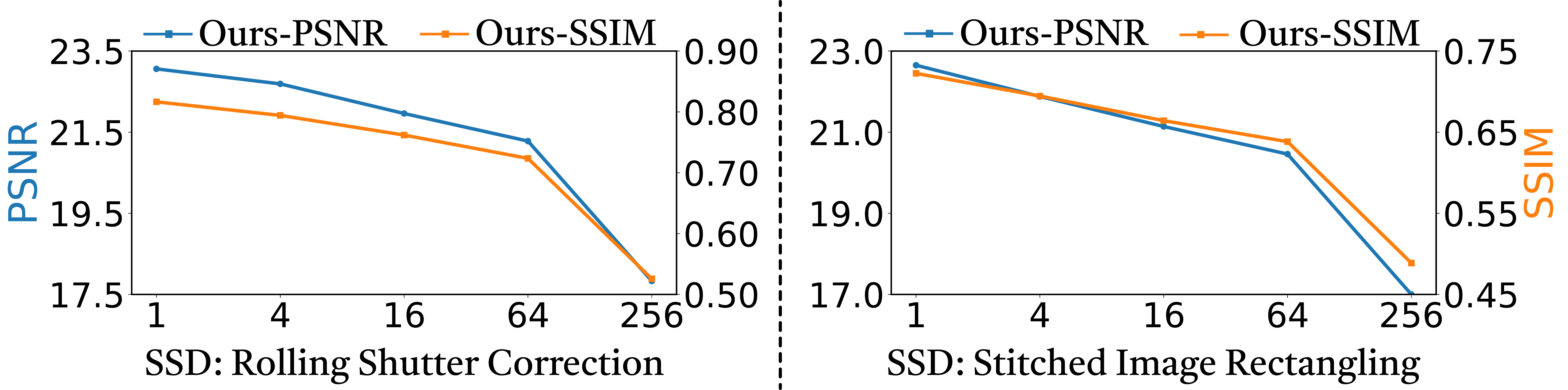}
    \caption{Quantitative comparison of different sampling steps.}
    \label{fig:sampling_steps_quali_}
\end{figure}

To investigate the impact of sampling steps on the final reconstruction quality, we conduct a controlled experiment using a DDIM scheduler~\cite{song2020denoising} with varying numbers of inference steps. All samples are generated from the same model checkpoint to ensure consistency.

The quantitative results are shown in Fig.~\ref{fig:sampling_steps_quali_}, where both PSNR and SSIM are reported across different step settings. A clear degradation trend is observed: as the number of inference steps increases, the image quality deteriorates. This is consistent with our theoretical analysis in Section~\ref{sec:ssd}, which suggests that prolonged iterative refinement tends to amplify errors in the predicted flow, resulting in structurally inconsistent or distorted outputs.

Fig.~\ref{fig:sampling_steps_quali} further supports this observation with visual comparisons. When using fewer steps, the output image preserves sharper structures and more faithful local details. In contrast, increasing the number of sampling steps leads to over-smoothed textures, distorted geometry, and artifacts near semantic boundaries. These results highlight the counterintuitive phenomenon we term \textit{Sampling Steps Disaster}, where applying more inference steps, which is typically assumed to improve sample quality, can in fact degrade performance in models trained with multiple supervision signals.

\subsection{Ablation Studies}\label{sec:ablat}

\paragraph{Effect of prior knowledge}
\label{sec:inferenceSteps}
{To assess the contribution of pretrained diffusion priors, we train two ablation variants on DIR-D~\cite{nie2022deep}. In the main StableMotion model, the pretrained VAE is kept frozen and the UNet is initialized from the pretrained diffusion model before fine-tuning. In the VAE-only-prior variant, the pretrained VAE is retained while the UNet is randomly initialized and trained under the same setting. In the no-prior variant, both the VAE and UNet are randomly initialized and jointly trained. All other architectural choices, hyperparameters, and optimization settings are kept the same.}

{As shown in Table~\ref{table:step_compare}, the no-prior variant failed to yield meaningful convergence under this common training setup and plateaued after 8k steps (12.00$\to$11.95$\to$11.93), indicating that simply extending training did not improve performance in our setting. The VAE-only-prior variant converges much better and reaches 22.35\,dB at 24k steps, close to the full-prior model at the same checkpoint (22.41\,dB). This shows that the pretrained VAE prior is already important for in-distribution optimization on DIR-D.}

{However, comparable in-distribution PSNR does not imply comparable cross-scenario generalization. Table~\ref{table:unet_prior_ood} compares the full-prior model and the VAE-only-prior variant on the SIR-OOD benchmark. Removing the UNet prior reduces the overall score from 6.86 to 6.42, with the largest drop on rectangular boundary quality (8.84$\to$7.25) and a clear decrease in artifact control (5.60$\to$5.24). Thus, while the VAE prior largely restores DIR-D PSNR, the UNet prior remains important for \rev{improved} boundary reasoning and artifact suppression on unseen SIR scenarios\rev{, as suggested by these no-reference scores}.}

{We attribute these results to the complementary roles of the VAE and UNet priors. We also tested training the VAE from scratch as a separate first stage, but this stage did not converge to a usable latent space; the natural fallback of joint training likewise did not yield meaningful convergence. Unlike the original LDM setting, where the first-stage VAE is trained on very large-scale image data, no comparable large-scale flow-field corpus is available in our case, so the staged recipe is not practically viable for the no-prior setting. Therefore, in the no-prior variant we report the result obtained by randomly initializing both the VAE and UNet and jointly training them under the same overall training setting.}

\begin{table}[tb]
    \centering
    \def\temptablewidth{0.49\textwidth}
    {\rule{\temptablewidth}{1pt}}
    \begin{tabular*}{\temptablewidth}{@{\extracolsep{\fill}}cc|ccccc}
        \hline
        VAE & UNet & 2k& 4k & 8k & 16k & 24k \\
        \hline
        \checkmark & \checkmark & \textbf{20.18} & \textbf{21.26} & \textbf{21.87} & \textbf{22.19} & \textbf{22.41}\\
        \hline
        {\checkmark} & & {18.03} & {19.58} & {21.05} & {22.06} & {22.35}\\
        \hline
        & & {11.06} & {11.88} & {12.00} & {11.95} & {11.93}\\
        \hline
    \end{tabular*}
    {\rule{\temptablewidth}{1pt}}
    \caption{{Ablation of diffusion priors, trained on DIR-D~\cite{nie2022deep}. The VAE and UNet columns indicate whether the corresponding pretrained prior is used. Columns (2k, 4k, \ldots, 24k) denote model checkpoints saved at the corresponding number of training steps. The metric reported is PSNR (dB).}}
    \label{table:step_compare}
\end{table}

\begin{table}[tb]
    \centering
    \def\temptablewidth{0.49\textwidth}
    {\color{black}
        {\rule{\temptablewidth}{1pt}}
        \scriptsize
        \begin{tabular*}{\temptablewidth}{@{\extracolsep{\fill}}lcccccc}
            \hline
            Method & Overall & Spec. & Geom. & Cont. & Art. & Nat.\\
            \hline
            w/o UNet Prior & 6.42 & 7.25 & \textbf{5.25} & 7.73 & 5.24 & 5.70\\
            w/ UNet Prior & \textbf{6.86} & \textbf{8.84} & 5.18 & \textbf{7.74} & \textbf{5.60} & \textbf{5.78}\\
            \hline
        \end{tabular*}
        {\rule{\temptablewidth}{1pt}}
    }
    \caption{{SIR-OOD generalization comparison for the UNet-prior ablation. Both variants use the pretrained VAE prior and are trained on DIR-D~\cite{nie2022deep}; the w/o UNet Prior variant randomly initializes the UNet. Spec. denotes rectangular boundary quality. Scores are assigned by GPT-5.4 from 0 to 10 and averaged over the SIR-OOD benchmark; higher is better.}}
    \label{table:unet_prior_ood}
\end{table}

\paragraph{Loss functions}
\begin{table}[htb]
    \centering
    \def\temptablewidth{0.46\textwidth}
    {\rule{\temptablewidth}{1pt}}
    \begin{tabular*}{\temptablewidth}{@{\extracolsep{\fill}}cccccc}
        \hline
        $\ell_{diff}$ & $\ell_{cond}$ & $\rev{\ell_{pct}}$ & PSNR$\uparrow$ & SSIM$\uparrow$ & TOPIQ$\uparrow$\\
        \hline
        \checkmark &  &  & 23.37 & 0.7957 & 0.7912\\
        \hline
        \checkmark & \checkmark & & 25.28 & 0.8351 & 0.8427\\
        \hline
        \checkmark & \checkmark & \checkmark & 25.50 & 0.8416 & 0.8511\\
        \hline
    \end{tabular*}
    {\rule{\temptablewidth}{1pt}}
    \caption{{Ablation on loss components. Trained and evaluated on RS-Real~\cite{yang2024single} at $256 \times 256$ resolution with ensemble $n{=}1$.}}
    \label{table:ablation_loss}
\end{table}

\begin{table}[htb]
    \centering
    \def\temptablewidth{0.46\textwidth}
    {\rule{\temptablewidth}{1pt}}
    \begin{tabular*}{\temptablewidth}{@{\extracolsep{\fill}}lcccc}
        \hline
        Ensemble num & 1 & 2 & 4 & 8\\
        \hline
        PSNR & {22.86} & {23.06} & {23.13} & {23.17} \\
        \hline
    \end{tabular*}
    {\rule{\temptablewidth}{1pt}}
    \caption{{Ablation of ensemble counts on the image rectangling task, trained \rev{on DIR-D}~\cite{nie2022deep}. The main result in Table~\ref{table:IR_metrics} uses $n{=}2$ (PSNR 23.06).}}
    \label{table:exp_abla_1}
\end{table}

To evaluate the individual contributions of the condition loss $\ell_{cond}$ and the perceptual loss $\rev{\ell_{pct}}$, we conducted an ablation study by training two additional variants of our model. Quantitative results are summarized in Table~\ref{table:ablation_loss}. The comparison highlights how each component contributes to the overall performance. All evaluations are conducted on the RS-Real dataset~\cite{yang2024single} at a resolution of $256 \times 256$.

\begin{figure}[htb]
    \vspace{-1em}
    \begin{center}
        \includegraphics[width=\linewidth]{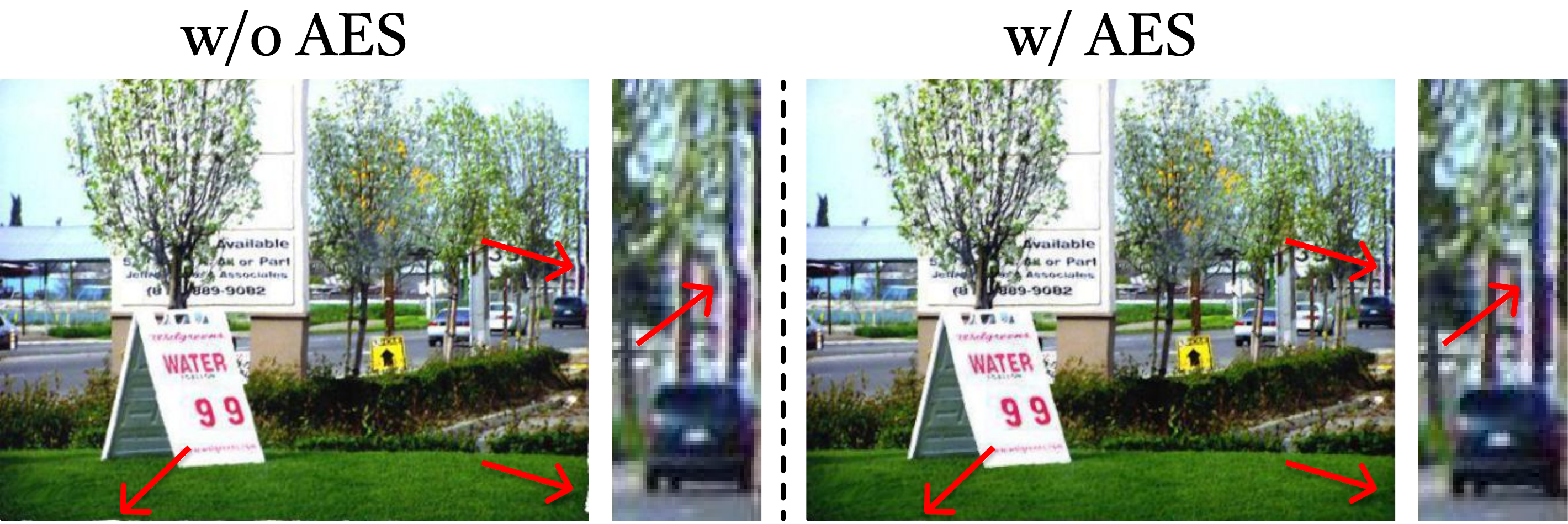}
    \end{center}
    \vspace{-1em}
    \caption{Distortions and boundaries are further repaired by AES.}
    \label{fig:ensemble}
\end{figure}

\paragraph{Adaptive ensemble strategy} The impact of the ensemble count is reported in Table~\ref{table:exp_abla_1}, and visual results are shown in Fig.~\ref{fig:ensemble}. AES reduces the uncertainty inherent in diffusion models and addresses unstable local regions as well as irregular boundaries in the image.

\paragraph{Image-to-image framework}

\begin{figure}[htb]
    \vspace{-1em}
    \begin{center}
        \includegraphics[width=\linewidth]{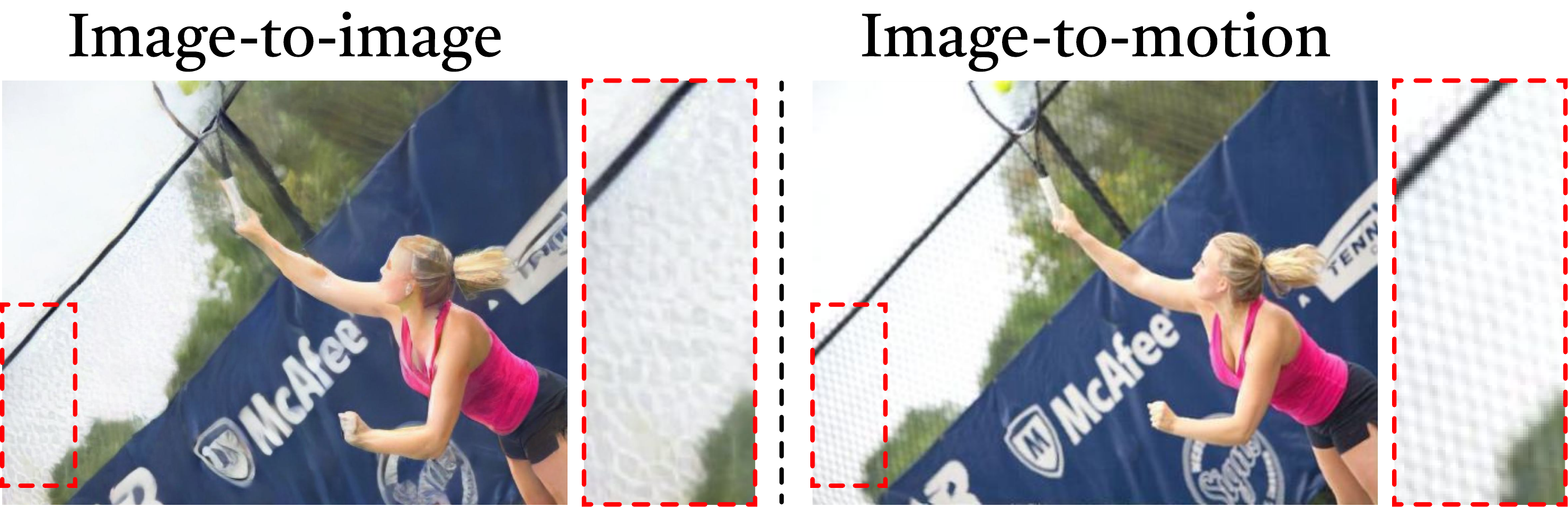}
    \end{center}
    \vspace{-1em}
    \caption{Compare StableMotion to image-to-image framework.}
    \label{fig:i2i}
\end{figure}

We also explored the image-to-image fine-tuning paradigm using Stable Diffusion (SD) on the DIR-D dataset~\cite{nie2022deep}. Despite convergence during training, the resulting model exhibited subpar performance, achieving a PSNR of only 19.73. Representative visual results are shown in Fig.~\ref{fig:i2i}. We attribute this degradation to the inherently generative nature of SD, which tends to produce locally inconsistent details in the absence of strong structural constraints. These unstable textures and spatial variations undermine geometric fidelity, ultimately compromising the model’s ability to accurately reconstruct rectified images.

\section{Conclusion}
In this work, we present StableMotion, a framework that leverages image priors for motion estimation and evaluate it on two tasks. Unlike previous image-prior-based image-to-image methods and diffusion-based methods trained from scratch on task-specific data with substantial compute, our method repurposes a text-to-image pretrained model (Stable Diffusion) into an image-to-motion framework, performs one-step inference, and achieves strong performance. We also introduce Sampling Steps Disaster (SSD), which characterizes a hidden issue in diffusion models trained with multiple learning objectives. SSD supports the one-step inference design of StableMotion and may be relevant to other diffusion-based tasks. We further propose Adaptive Ensemble Strategy (AES) to reduce output variability. Overall, StableMotion outperforms previous methods on the public benchmarks we evaluate, and {the qualitative zero-shot examples together with no-reference evaluations on unseen SIR-OOD and real-captured RSC benchmarks suggest promising transfer to unseen data}.

{\textbf{Limitations and future work.} \rev{Our reported timings are measured on an NVIDIA H100 and therefore demonstrate the \emph{algorithmic} efficiency of StableMotion relative to prior multi-step diffusion baselines, since it avoids the iterative denoising loop. However, they do not by themselves establish deployment performance on mobile NPUs or consumer-grade devices.} While the one-step design is inherently more amenable to resource-constrained deployment than multi-step diffusion, evaluating \rev{on-device latency and memory on} edge devices and mobile NPUs through model compression techniques (e.g., quantization, pruning) remains an important direction for future work.}

\bibliographystyle{IEEEtran}
\bibliography{main}

\begin{IEEEbiography}
    [{\includegraphics[width=1in,height=1.25in,clip,keepaspectratio]{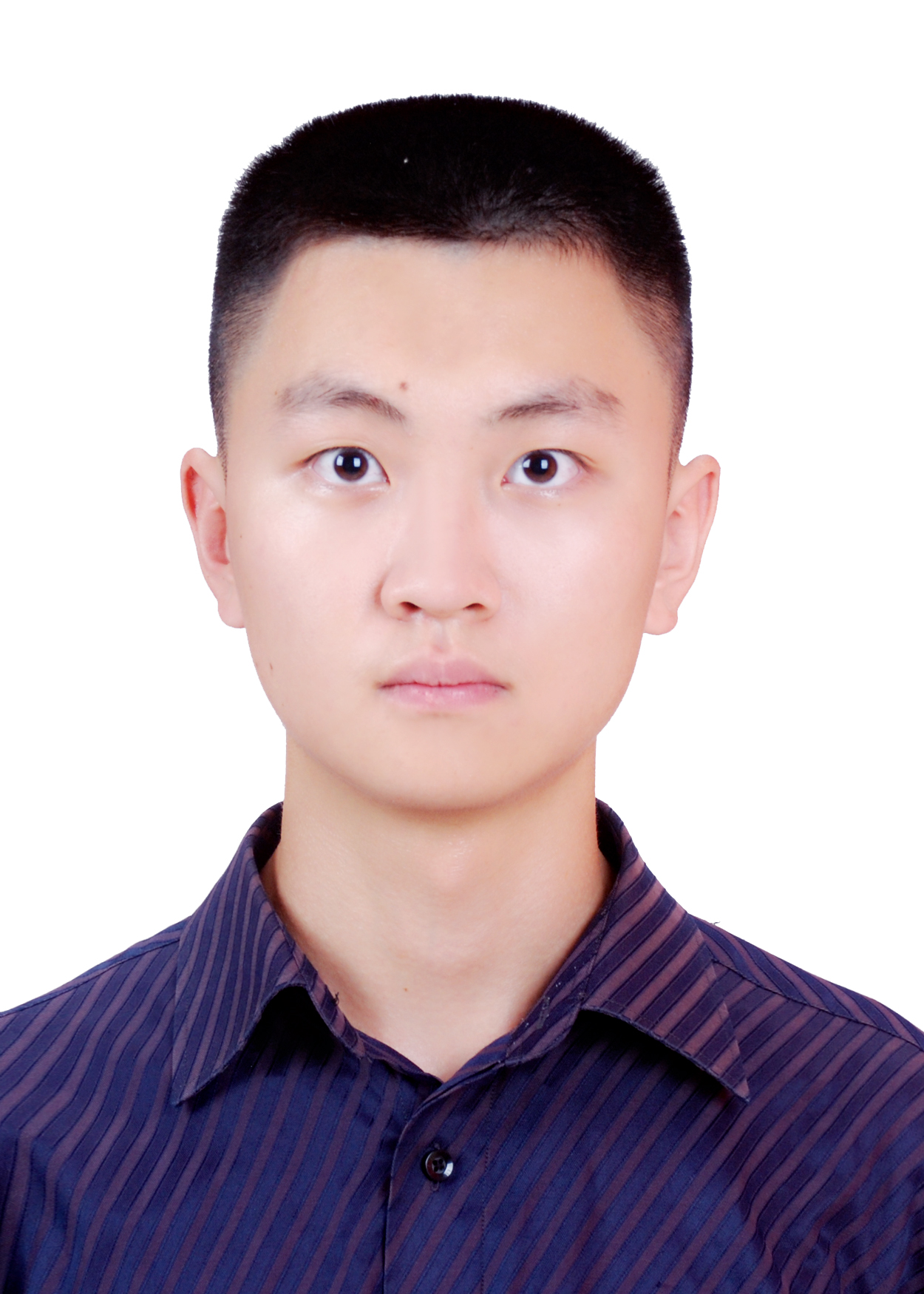}}]{Ziyi Wang} is currently pursuing the B.S. degree at the Yingcai Honors College, University of Electronic Science and Technology of China. His research interests include computer vision, computational photography and generative models.
\end{IEEEbiography}

\begin{IEEEbiography}
    [{\includegraphics[width=1in,height=1.25in,clip,keepaspectratio]{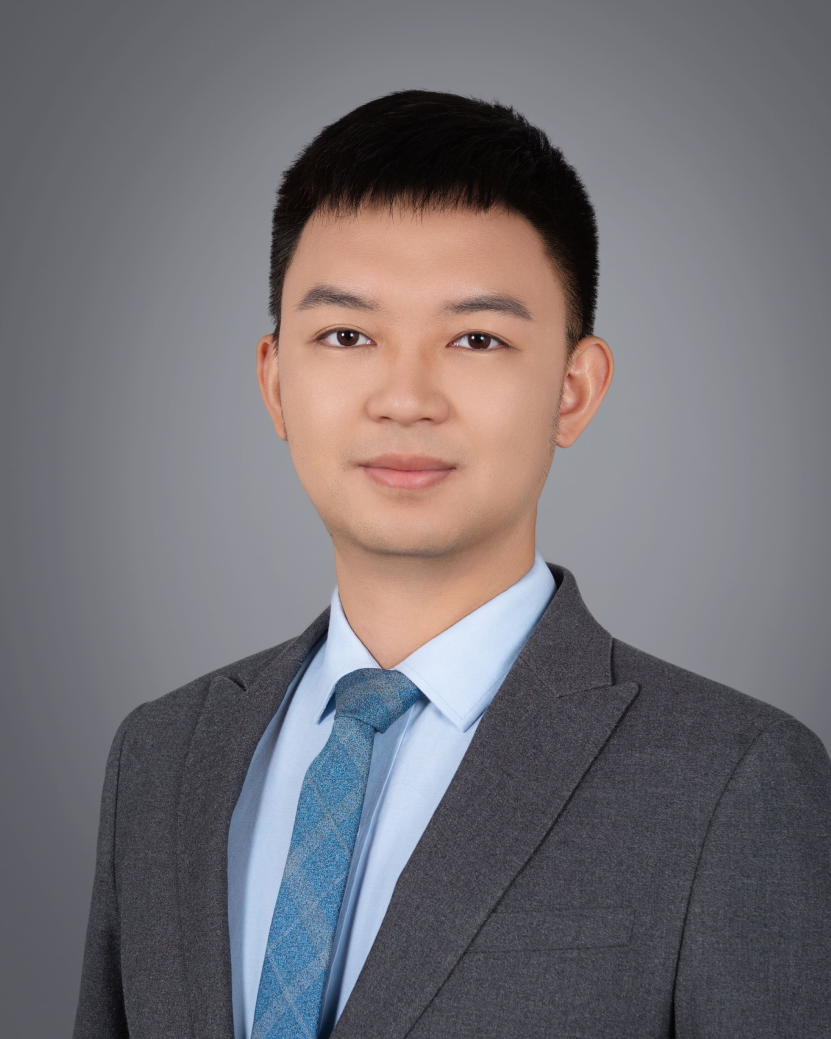}}]{Haipeng Li} received the BEng degree from the University of Electronic Science and Technology of China (UESTC), Chengdu, China, in 2017 and the MSc degree from the Institut Mines-Telecom Atlantique Bretagne Pays de la Loire, Brest, France, in 2020. He was a Researcher in Megvii Research Chengdu during 2019-2022. He has been a PhD student since 2022 in the School of Information and Communication Engineering, UESTC. His research interests include generative models and computer vision. In the past three years, he has published several papers in top journals and conferences.
\end{IEEEbiography}

\begin{IEEEbiography}
    [{\includegraphics[width=1in,height=1.25in,clip,keepaspectratio]{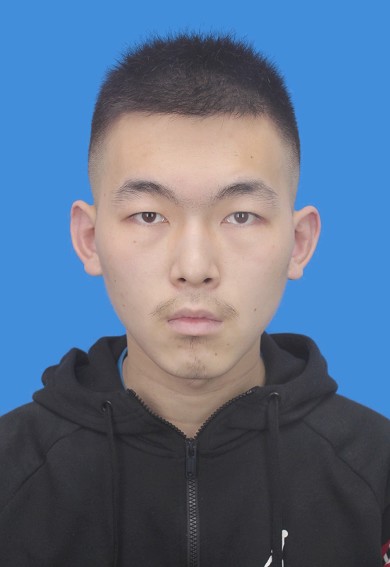}}]{Lin Sui} received the BEng degree in Software Engineering from Xi’an Jiaotong University (XJTU), Xi'an, China, in 2020, and the MSc degree in Computer Science and Technology from Nanjing University (NJU), Nanjing, China, in 2023. He is currently a computer vision researcher in the 4Paradigm Inc., Beijing, China. His research interests include computer vision and machine learning. In the past three years, he has published several papers in top journals and conferences such as IEEE TPAMI, CVPR, ECCV, and WACV.
\end{IEEEbiography}

\begin{IEEEbiography}
    [{\includegraphics[width=1in,height=1.25in,clip,keepaspectratio]{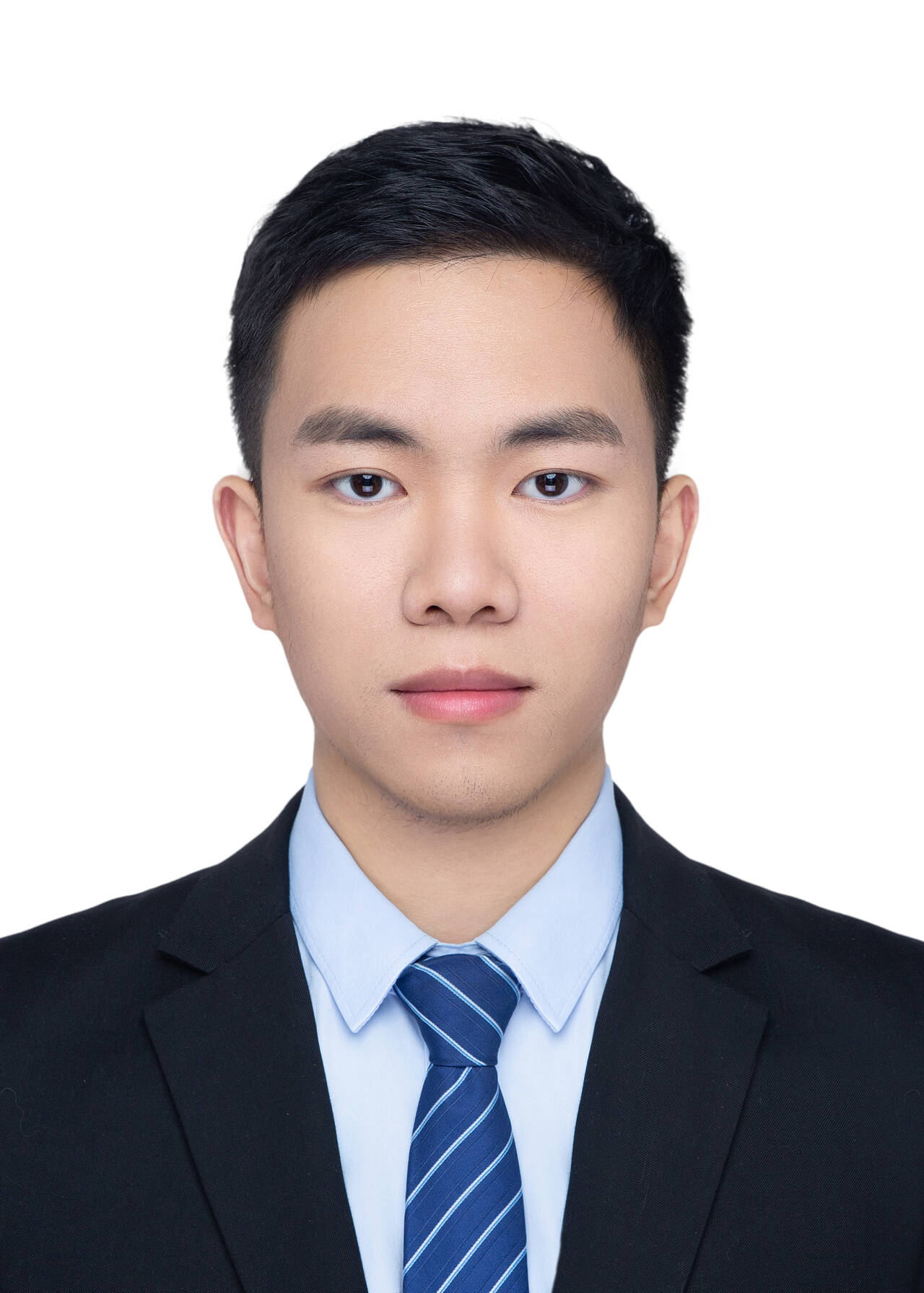}}]{Tianhao Zhou} is an undergraduate student in Yingcai Honors College of the University of Electronic Science and Technology of China. His current research \rev{interests include} computer vision, deep learning and generative models.
\end{IEEEbiography}

\begin{IEEEbiography}
    [{\includegraphics[width=1in,height=1.25in,clip,keepaspectratio]{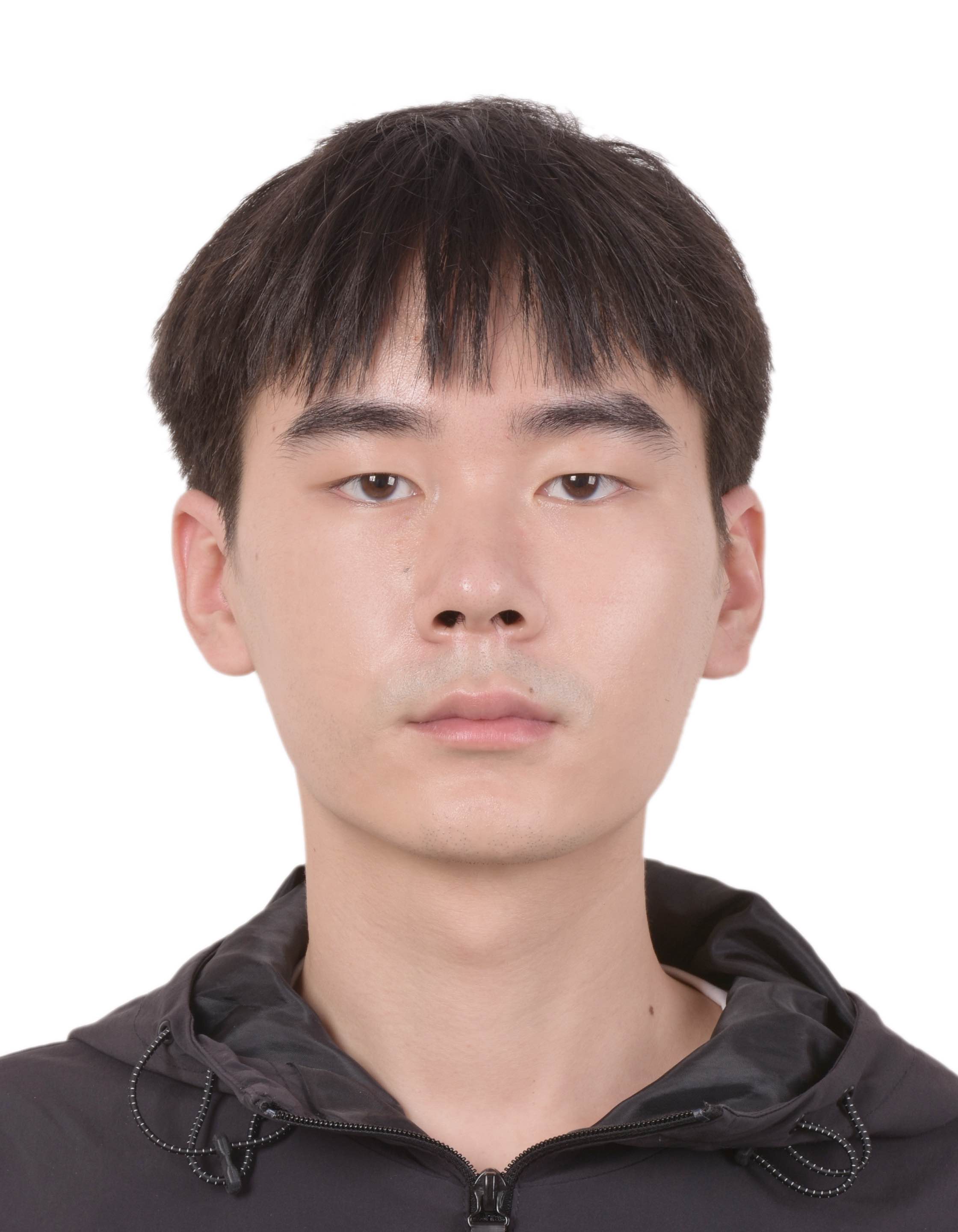}}]{Hai Jiang} received the BEng degree and MSc degree from the School of Aeronautics and Astronautics, Sichuan University, Chengdu, China, in 2016 and 2020, respectively. He is currently working toward the PhD degree with the School of Aeronautics and Astronautics, Sichuan University, and an intern with Megvii Technology. His research interests include computer vision and deep learning.
\end{IEEEbiography}

\begin{IEEEbiography}
    [{\includegraphics[width=1in,height=1.25in,clip,keepaspectratio]{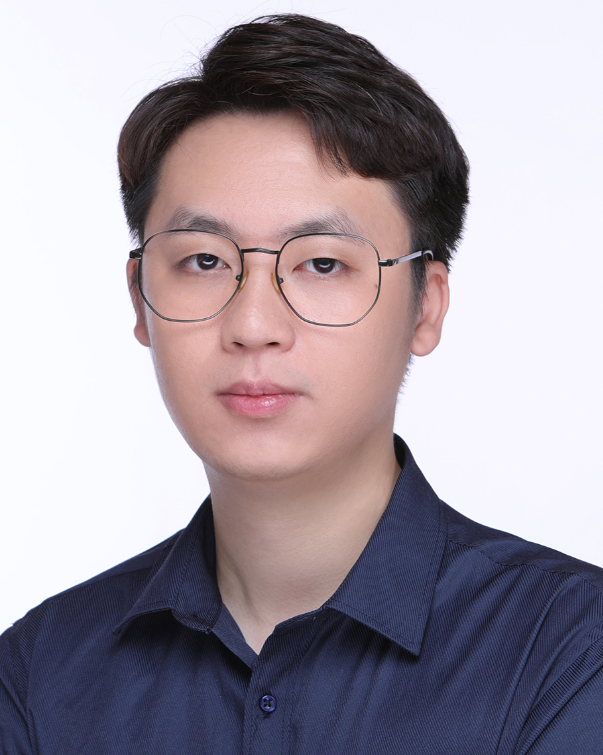}}]{Lang Nie} is  currently  an  Associate  Professor  at Chongqing University of Posts and Telecommunications, China. Prior to that, he received his B.S. and Ph.D. degrees from the School of Computer Science and Technology, Beijing Jiaotong University (BJTU) in  2019  and  2025. His current research interests include image and video processing, 3-D vision, and multi-view geometry.
\end{IEEEbiography}

\begin{IEEEbiography}[{\includegraphics[width=1in,height=1.25in,clip,keepaspectratio]{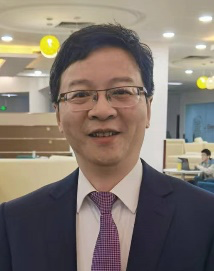}}]{Bing Zeng}
    (M'91-SM'13-F'16) received the BEng and MEng degrees in electronic engineering from University of Electronic Science and Technology of China (UESTC), Chengdu, China, in 1983 and 1986, respectively, and the PhD degree in electrical engineering from Tampere University of Technology, Tampere, Finland, in 1991.

    He worked as a postdoctoral fellow at University of Toronto from September 1991 to July 1992 and as a Researcher at Concordia University from August 1992 to January 1993. He then joined the Hong Kong University of Science and Technology (HKUST). After 20 years of service at HKUST, he returned to UESTC in the summer of 2013. At UESTC, he leads the Institute of Image Processing to work on image and video processing, multimedia communication, computer vision, and AI technology, was Dean of Glasgow College (a joint school between UESTC and University of Glasgow) during 2018-2022, and has been a Vice Chair of Committee for Academic Affairs since 2016.

    He served as an Associate Editor for IEEE TCSVT for 8 years and received the Best Associate Editor Award in 2011. He was General Co-Chair of IEEE VCIP-2016 and PCM-2017. He received a 2nd-Class Natural Science Award (the 1st recipient) from Chinese Ministry of Education in 2014 and was elected as an IEEE Fellow in 2016 for contributions to image and video coding.
\end{IEEEbiography}

\begin{IEEEbiography}[{\includegraphics[width=1in,height=1.25in,clip,keepaspectratio]{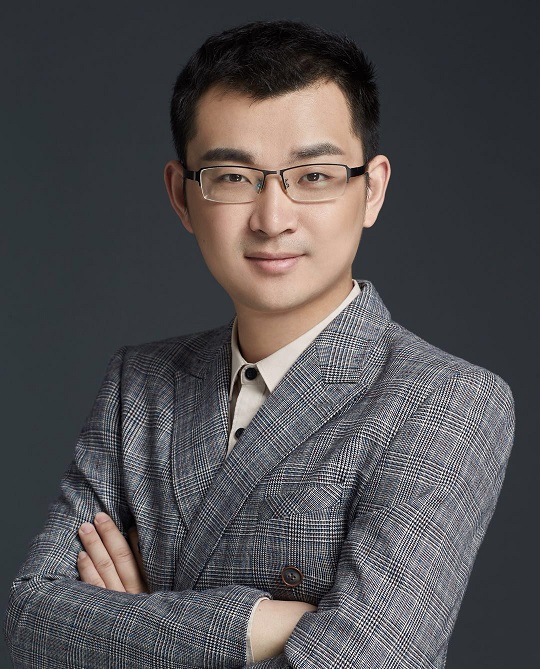}}]{Shuaicheng Liu} (M'14-SM'23) received the Ph.D. and M.Sc. degrees from the National University of Singapore, Singapore, in 2014 and 2010, respectively, and the B.E. degree from Sichuan University, Chengdu, China, in 2008. In 2015, he joined the University of Electronic Science and Technology of China and is currently a Professor with the Institute of Image Processing, School of Information and Communication Engineering, Chengdu, China. His research interests include computer vision and computer graphics.
\end{IEEEbiography}



\end{document}